\documentclass[table]{article}
\PassOptionsToPackage{table}{xcolor}

\usepackage{colm2024_conference}

\usepackage{booktabs}
\usepackage{graphicx}
\usepackage{graphics}
\usepackage{enumitem}
\usepackage{wrapfig}
\usepackage{algorithm}
\usepackage{algpseudocode}
\usepackage{multicol}
\usepackage{hyperref}   
\usepackage{xcolor}

\usepackage{microtype}
\usepackage{amsmath}
\usepackage{colortbl}
\usepackage[utf8]{inputenc}
\definecolor{lightgray}{rgb}{0.9,0.9,0.9}
\usepackage{caption}
\usepackage{subcaption}
\usepackage{xcolor}
\usepackage{setspace}
\usepackage{url}
\usepackage{multirow}
\usepackage{colortbl}
\usepackage{tabularx}
\usepackage{blindtext}
\usepackage{pgfplots}
\pgfplotsset{compat=1.18} 
\usepackage{tikz}
\usetikzlibrary{er,positioning,bayesnet}
\usepackage{makecell}
\usepackage{tipa}
\usepackage{siunitx}
\usepackage{nicefrac}
\usepackage{tocloft}
\usepackage{listings}
\usepackage[raster,skins]{tcolorbox} %
\usepackage{xltabular}
\usepackage{adjustbox}
\usepackage{xurl}
\usepackage{setspace}

\usepackage{lipsum}
\usepackage{ulem}

\usepackage{arydshln} 
\usepackage[table]{xcolor}
\usepackage{varwidth}

\usepackage{amsmath,amsfonts,bm}

\def\eqref#1{equation~\ref{#1}}

\def\1{\bm{1}}

\DeclareMathAlphabet{\mathsfit}{\encodingdefault}{\sfdefault}{m}{sl}
\SetMathAlphabet{\mathsfit}{bold}{\encodingdefault}{\sfdefault}{bx}{n}

\title{AutoCodeBench: Large Language Models are Automatic Code Benchmark Generators}

\newcommand*\samethanks[1][\value{footnote}]{\footnotemark[#1]}

\author{
    \textbf{Jason Chou}\thanks{Equal Contributions.}
    \quad
    \textbf{Ao Liu}\samethanks
    \quad
    \textbf{Yuchi Deng}
    \quad
    \textbf{Zhiying Zeng}
    \quad
    \textbf{Tao Zhang}
    \quad
    \textbf{Haotian Zhu}
    \quad
    \textbf{Jianwei Cai}
    \quad
    \textbf{Yue Mao}
    \quad
    \textbf{Chenchen Zhang}
    \quad
    \textbf{Lingyun Tan}
    \quad
    \textbf{Ziyan Xu}
    \quad
    \textbf{Bohui Zhai}
    \quad
    \textbf{Hengyi Liu}
    \quad
    \textbf{Speed Zhu}
    \quad
    \textbf{Wiggin Zhou}~\thanks{Corresponding Authors.}
    \quad
    \textbf{Fengzong Lian}~\samethanks
    \quad
    \vspace{.5em}\\
    Hunyuan Team, Tencent
    \vspace{.5em}\\
    \texttt{\{wigginzhou,faxonlian\}@tencent.com}
    \vspace{.5em}\\
    \includegraphics[width=0.35cm]{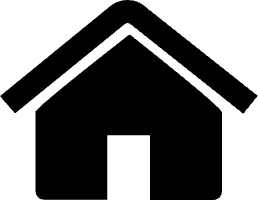} \ \ 
    \textcolor{blue!60!black}{\href{https://autocodebench.github.io/}{Homepage}}
}

\begin{document}

\maketitle

\begin{abstract}
Large Language Models (LLMs) have demonstrated remarkable capabilities across various domains, with code generation emerging as a key area of focus. While numerous benchmarks have been proposed to evaluate their code generation abilities, these benchmarks face several critical limitations. First, they often rely on manual annotations, which are time-consuming and difficult to scale across different programming languages and problem complexities. Second, most existing benchmarks focus primarily on Python, while the few multilingual benchmarks suffer from limited difficulty and uneven language distribution.
To address these challenges, we propose \textbf{AutoCodeGen}, an automated method for generating high-difficulty multilingual code generation datasets without manual annotations. AutoCodeGen ensures the correctness and completeness of test cases by generating test inputs with LLMs and obtaining test outputs through a multilingual sandbox, while achieving high data quality through reverse-order problem generation and multiple filtering steps.
Using this novel method, we introduce \textbf{AutoCodeBench}, a large-scale code generation benchmark comprising 3,920 problems evenly distributed across 20 programming languages. It is specifically designed to evaluate LLMs on challenging, diverse, and practical multilingual tasks. We evaluate over 30 leading open-source and proprietary LLMs on AutoCodeBench and its simplified version \textbf{AutoCodeBench-Lite}. The results show that even the most advanced LLMs struggle with the complexity, diversity, and multilingual nature of these tasks. Besides, we introduce \textbf{AutoCodeBench-Complete}, specifically designed for base models to assess their few-shot code generation capabilities. We hope the AutoCodeBench series will serve as a valuable resource and inspire the community to focus on more challenging and practical multilingual code generation scenarios.


\end{abstract}




\section{Introduction}
\label{sec:intro}



Recently, Large Language Models (LLMs) have undergone rapid development, demonstrating impressive capabilities in various tasks~\citep{gpt4o,gemini2.5,deepseekai2025deepseekr1,tencenthunyuanteam2025hunyuanturbos}. Among these, code generation has emerged as a key indicator of both model intelligence and practical utility, attracting growing attention from both academia and industry~\citep{chen2021evaluatinglargelanguagemodels,jimenez2024swebench,jiang2024surveylargelanguagemodels}. By generating executable code, LLMs have the potential to significantly enhance programming automation and alleviate the burden of manual coding. Many popular and powerful LLMs like Claude 4~\citep{claude4} and DeepSeek-V3~\citep{deepseek_v3} have already been widely adopted in AI-assisted coding scenarios~\citep{Cursor}.

To measure and improve the code generation capabilities of LLMs, numerous code generation benchmarks are proposed~\citep{wang2025softwaredevelopmentlifecycle}. Early works such as HumanEval~\citep{chen2021evaluatinglargelanguagemodels} and MBPP~\citep{austin2021programsynthesislargelanguage} laid the foundation by evaluating LLMs' abilities on short, algorithm-focused Python problems. However, these simple and narrowly defined tasks have quickly become outdated with the rapid development in LLMs. Recent benchmarks~\citep{quan2025codeelobenchmarkingcompetitionlevelcode,jain2025livecodebench,zheng2025livecodebenchproolympiadmedalists,zhu2025oibenchbenchmarkingstrongreasoning} have shifted focus to more challenging, competition-level Python problems with the help of professional manual annotations. 
On the other hand, some other benchmarks~\citep{peng2024humanevalxlmultilingualcodegeneration,chai2025mceval,fullstackbench}, such as FullStackBench~\citep{fullstackbench}, emphasize more practical multilingual programming scenarios. However, manually crafting these programming problems and test cases is time-consuming and labor-intensive, making it difficult to simultaneously achieve high difficulty and multilingual coverage, as shown in Table~\ref{introduction_benchmark}. \textbf{\textit{Therefore, the community currently needs a benchmark that combines high difficulty, practical diversity, and balanced multilingual distribution to comprehensively evaluate the code generation capabilities of LLMs.}}

\begin{table}[]
\small
\resizebox{\textwidth}{!}{
\begin{tabular}{lcccccccc}
\toprule
\textbf{Benchmark} & \textbf{MLing} & \textbf{MLogi} & \textbf{HFree} & \textbf{BDist} & \textbf{Difficulty} & \textbf{Category} & \textbf{Data Size} & \textbf{Problem Len}  \\ \midrule
HumanEval          & \includegraphics[width=0.35cm]{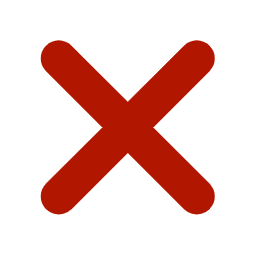}                & \includegraphics[width=0.35cm]{./figures/no.png}                  & \includegraphics[width=0.35cm]{./figures/no.png}              & /    & \includegraphics[width=0.25cm]{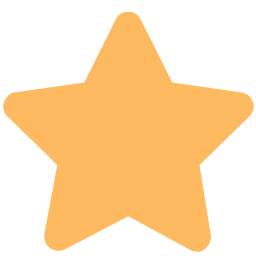}                    & \textcolor{gray!80!black}{5} & \textcolor{gray!80!black}{164}                & \textcolor{orange!80!black}{134.1}                    \\
MBPP               & \includegraphics[width=0.35cm]{./figures/no.png}                    & \includegraphics[width=0.35cm]{./figures/no.png}                  & \includegraphics[width=0.35cm]{./figures/no.png}               & /    & \includegraphics[width=0.25cm]{./figures/star.png}                   &  \textcolor{gray!80!black}{6} & \textcolor{gray!80!black}{378}                & \textcolor{gray!80!black}{50.5}                     \\
BigCodeBench       & \includegraphics[width=0.35cm]{./figures/no.png}                    & \includegraphics[width=0.35cm]{./figures/no.png}                  & \includegraphics[width=0.35cm]{./figures/no.png}                & /    & \includegraphics[width=0.25cm]{./figures/star.png}\includegraphics[width=0.25cm]{./figures/star.png}\includegraphics[width=0.25cm]{./figures/star.png}                 &  \textcolor{orange!80!black}{8} & \textcolor{orange!80!black}{1140}               & \textcolor{orange!80!black}{146.7}                    \\
LiveCodeBench      & \includegraphics[width=0.35cm]{./figures/no.png}                    & \includegraphics[width=0.35cm]{./figures/no.png}                  & \includegraphics[width=0.35cm]{./figures/no.png}               &  /     & \includegraphics[width=0.25cm]{./figures/star.png}\includegraphics[width=0.25cm]{./figures/star.png}\includegraphics[width=0.25cm]{./figures/star.png}\includegraphics[width=0.25cm]{./figures/star.png}\includegraphics[width=0.25cm]{./figures/star.png}               &  \textcolor{gray!80!black}{4} & \textcolor{orange!80!black}{1100}               & \textcolor{green!80!black}{469.6}                    \\
FullStackBench     & \includegraphics[width=0.35cm]{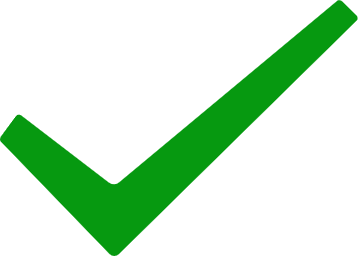}                   & \includegraphics[width=0.35cm]{./figures/no.png}                  & \includegraphics[width=0.35cm]{./figures/no.png}              &   \includegraphics[width=0.35cm]{./figures/no.png}    & \includegraphics[width=0.25cm]{./figures/star.png}\includegraphics[width=0.25cm]{./figures/star.png}                  &  \textcolor{green!80!black}{12} & \textcolor{orange!80!black}{1687}               & \textcolor{orange!80!black}{184.3}                    \\
McEval             & \includegraphics[width=0.35cm]{./figures/yes.png}                   & \includegraphics[width=0.35cm]{./figures/no.png}                  & \includegraphics[width=0.35cm]{./figures/no.png}         & \includegraphics[width=0.35cm]{./figures/yes.png}         & \includegraphics[width=0.25cm]{./figures/star.png}\includegraphics[width=0.25cm]{./figures/star.png}                  &  \textcolor{orange!80!black}{9} & \textcolor{green!80!black}{2007}               & \textcolor{orange!80!black}{146.7}                    \\ \midrule
\textbf{AutoCodeBench}      & \includegraphics[width=0.35cm]{./figures/yes.png}                   & \includegraphics[width=0.35cm]{./figures/yes.png}                 & \includegraphics[width=0.35cm]{./figures/yes.png}             & \includegraphics[width=0.35cm]{./figures/yes.png}    & \includegraphics[width=0.25cm]{./figures/star.png}\includegraphics[width=0.25cm]{./figures/star.png}\includegraphics[width=0.25cm]{./figures/star.png}\includegraphics[width=0.25cm]{./figures/star.png}\includegraphics[width=0.25cm]{./figures/star.png}               &  \textbf{\textcolor{green!80!black}{14}} & \textbf{\textcolor{green!80!black}{3920}}               & \textbf{\textcolor{green!80!black}{498.2}} \\ 
\bottomrule
\end{tabular}
}
\caption{Comparison of Existing Code Generation Benchmarks and AutoCodeBench. \textbf{MLing}: MultiLingual; \textbf{MLogi}: MultiLogical, refers to programming problems that require the model to simultaneously implement multiple core functionalities. \textbf{HFree}: Human-Free; \textbf{BDist}: Balanced Distribution of multiple languages. The \textbf{Difficulty} is rated based on the performance of \texttt{DeepSeek-V3-0324}. The number of \textbf{Category} is obtained using predefined labels from \texttt{DeepSeek-V3-0324}. The \textbf{Problem Len}gth is calculated using \texttt{Qwen2.5-32B-Instruct} tokenizer.}
\label{introduction_benchmark}
\end{table}

In this paper, we propose \textbf{AutoCodeGen}, an automated workflow centered around LLM-Sandbox Interaction, to accurately synthesize high-difficulty multilingual code generation datasets without any manual annotations. Unlike previous data synthesis strategies~\citep{luo2024wizardcoder,magicoder,xu2025kodcode,li2025codeio}, we ensure high data quality by constructing multilingual sandbox to generate test outputs and synthesizing programming problems in reverse order. Concretely, AutoCodeGen consists of the following key steps: 1) \textbf{Solution Generation}: Based on real-world code snippets, LLMs evolve self-contained code solutions, ensuring their practicality and diversity. 2) \textbf{Test Function Generation}: LLMs generate test inputs, which are concatenated with the code solutions and executed in the sandbox to obtain the test outputs. The two are then combined to form a complete test function. Compared to directly generating complete test cases as in KodCode~\citep{xu2025kodcode} or using Input Generator~\citep{li2025codeio}, our method efficiently ensures the correctness and completeness of the test cases. 3) \textbf{Problem Generation}: LLMs generate challenging programming problems based on heuristic specifications and integrates public test cases into them. 4) \textbf{Filtering}: Finally, we filter the data by Multiple Sampling, LLM-as-Critic, and Tagging to maintain the high-difficulty, high-quality, and diversity. 

Based on the automation workflow mentioned above, we propose \textbf{AutoCodeBench}, a large-scale, human-free code generation benchmark featuring 3,920 problems, as shown in Table~\ref{introduction_benchmark}. Compared with previous multilingual benchmarks~\citep{cassano2022multiplescalableextensibleapproach,fullstackbench,chai2025mceval}, ours simultaneously offers high difficulty, diversity, and practicality, with a balanced distribution of problems across 20 programming languages. Besides, we intentionally include multi-logical problems to assess the LLMs' ability to handle multi-logical reasoning. We believe this capability is important in the code agents scenarios like SWE-Bench~\citep{jimenez2024swebench}. 

The key contributions of this paper are as follows:

\begin{enumerate}
    \item \textbf{AutoCodeGen}. We propose an automated workflow based on LLM-Sandbox Interaction, where LLMs generate test inputs and obtain test outputs through the sandbox, to create high-quality code generation benchmarks. Worth mentioning is that this workflow can also be applied to synthesize high-quality training data.
    \item \textbf{AutoCodeBench}. We introduce AutoCodeBench, a large-scale code generation benchmark with 3,920 problems, evenly distributed across 20 programming languages, featuring high-difficulty, practicality, and diversity. Based on the evaluation results, we construct a simplified version AutoCodeBench-Lite. Besides, We tailor AutoCodeBench-Complete, a completion-based code generation benchmark, to assess the performance of base models.
    \item \textbf{Multilingual Sandbox}. We open-source a sandbox that supports 20+ programming languages. It is capable of high concurrency and request-based calls, making it suitable for multilingual evaluation and large-scale training.
    \item \textbf{Experimental Results}. We evaluate 30+ open-source and proprietary models. The results show that even the most advanced LLMs still struggle with complex and diverse multilingual programming tasks, especially in multi-logical scenarios.
    \item \textbf{Data and Experiment Analysis}. We conduct a comprehensive analysis of AutoCodeGen and AutoCodeBench, focusing on key aspects such as the quality, difficulty, and diversity of the generated data, as well as potential model biases during the generation process. We hope that these insights can provide valuable experience for the community in the development of future code generation benchmarks.
\end{enumerate}










\section{AutoCodeBench: A Challenging, Practical, and Diverse Multilingual Benchmark}

In this section, we first provide an overall data statistics of the AutoCodeBench and its simplified version AutoCodeBench-Lite. Following this, we introduce the automated workflow, AutoCodeGen, used to generate AutoCodeBench(-Lite). 

\subsection{Data Overview}

As shown in Table~\ref{introduction_benchmark} and ~\ref{method_bench}, AutoCodeBench is a large-scale, high-difficulty multilingual benchmark. Over 60\% of the problems are classified as hard problems, with each problem averaging 498.2 characters and accompanied by 9.6 test cases, providing a challenging and comprehensive evaluation standard. The 20 languages are as follows: \textit{Python, C++, Java, JavaScript(JS), Go, Shell, C\#, Dart, Elixir, Julia, Kotlin, Perl, PHP, Racket, R, Ruby, Rust, Scala, Swift, TypeScript(TS)}. 

\begin{figure}[t]
\centering
\begin{subfigure}[b]{0.45\textwidth}
    \centering
    \includegraphics[width=\textwidth]{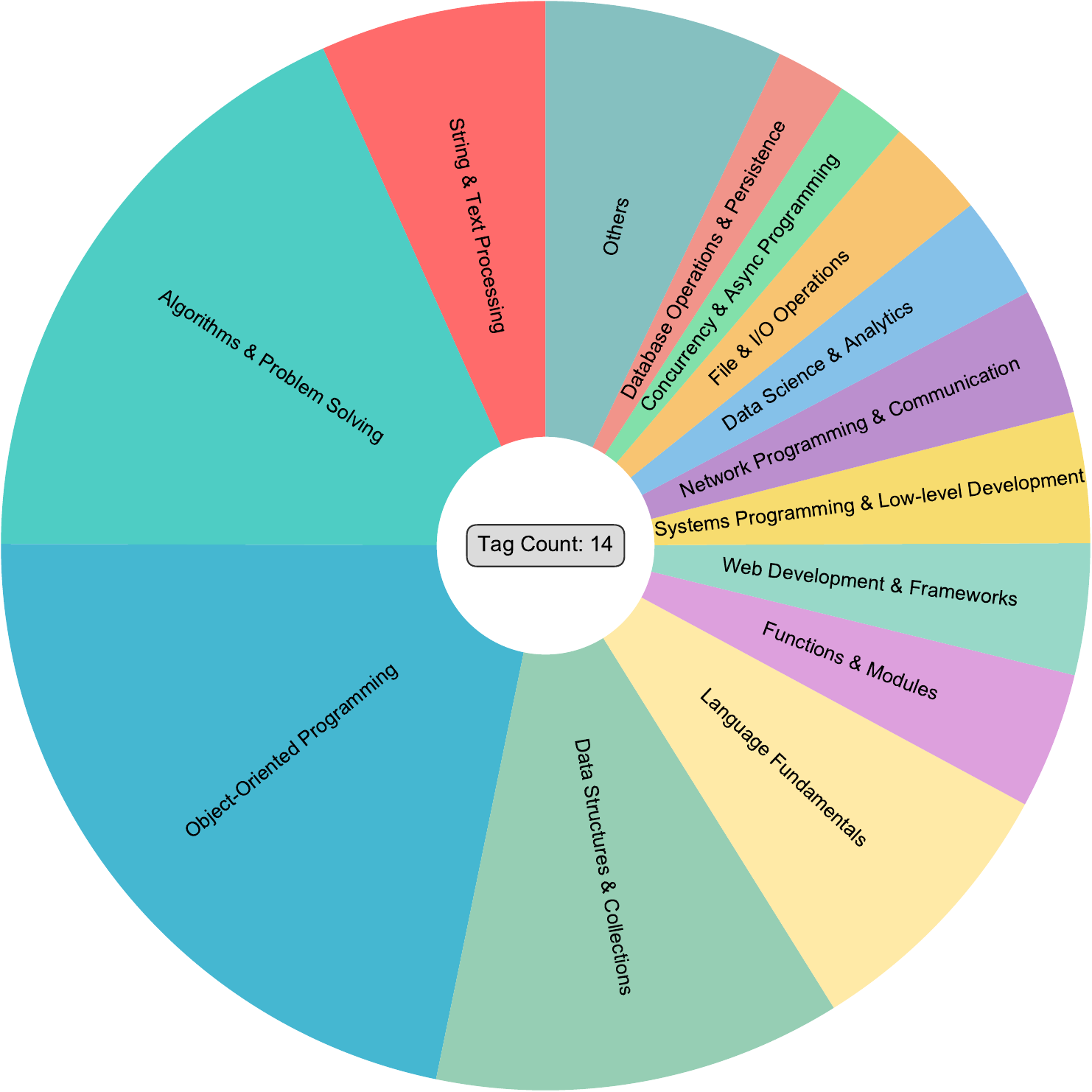}
    \label{fig:tag_acb}
\end{subfigure}
\hfill
\begin{subfigure}[b]{0.45\textwidth}
    \centering
    \includegraphics[width=\textwidth]{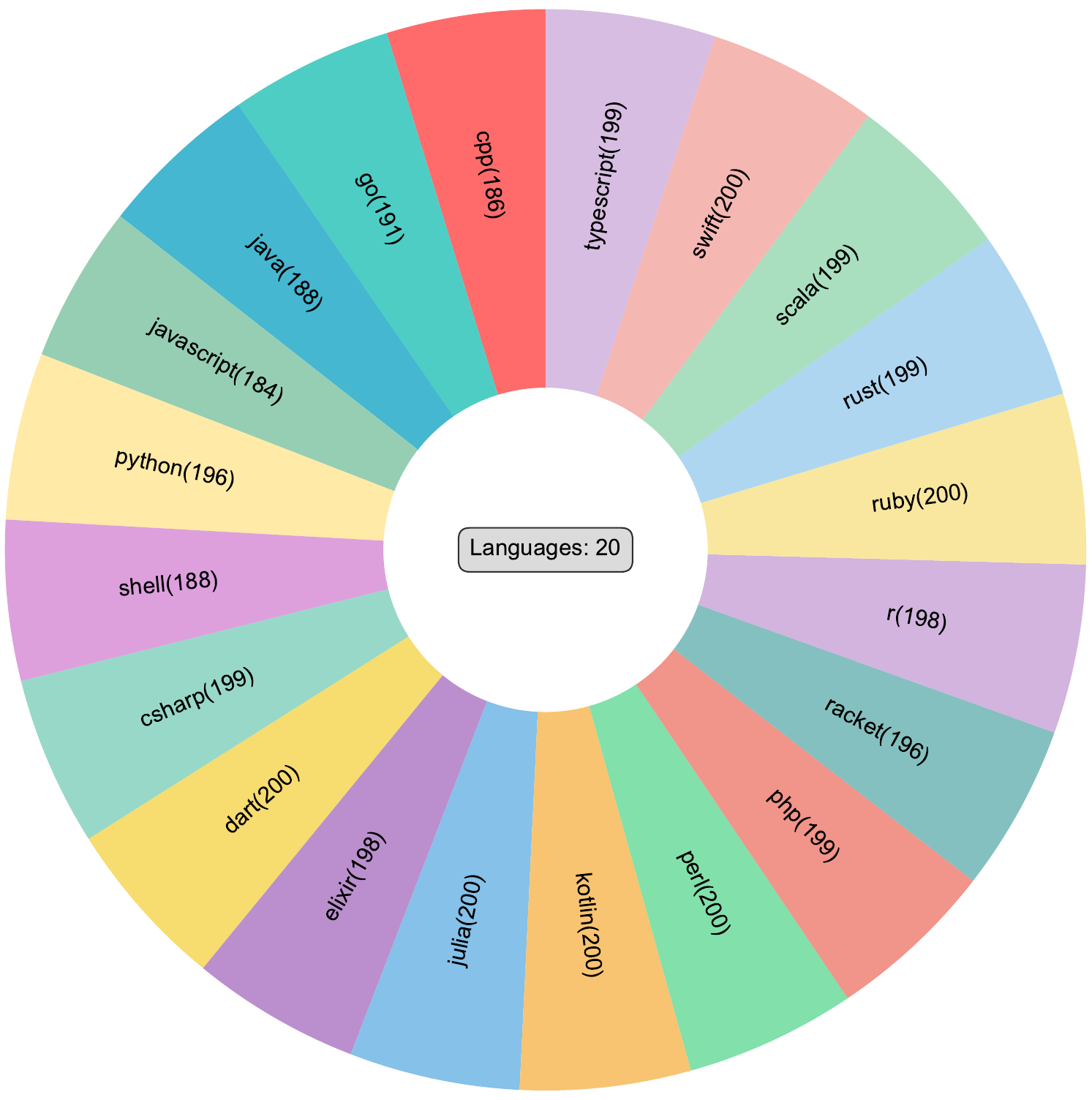}
    \label{fig:lan_acb}
\end{subfigure}

\caption{Tag and Language Distribution across our AutoCodeBench.}
\label{fig:statistic_acb}
\end{figure}

To analyze the diversity and language coverage of AutoCodeBench, we first use \texttt{Claude Sonnet 4} to generate 20 language-agnostic task categories, and then employ \texttt{DeepSeek-V3-0324} to classify each problem accordingly. Categories with less than 2\% representation are merged into the ``Other'' group. As shown in Figure~\ref{fig:statistic_acb}, AutoCodeBench covers 14 categories, demonstrating comprehensive coverage of practical programming scenarios. Besides, we analyze the distribution of problems across the 20 programming languages. AutoCodeBench exhibits a relatively balanced distribution across languages, with no significant bias toward any specific one, further validating its completeness and representativeness as a multilingual benchmark. Category labels and statistics of other benchmarks are provided in Appendix~\ref{appendix:a}.

\begin{table}[]
\begin{tabular}{lcccccc}
\toprule
                   & \#Problems & \#Test Cases & \#Langs & Prob Len  & Solu Len & Difficulty (Easy/Med/Hard) \\
\midrule
ACB      & 3,920      & 37,777       & 20          & 498.2                & 487.5                 & 646/846/2428                    \\
ACB-Lite &   1,586         &  15,341            &    20         &                     517.2         &                          469.3     & 263/421/902  \\
\bottomrule
\end{tabular}
\caption{Statistics of ACB and ACB-Lite. \textbf{ACB}: AutoCodeBench; \textbf{Langs}: Languages; \textbf{Prob}: Problem; \textbf{Solu}: Solution; \textbf{Len}: Length; \textbf{Med}: Medium. The difficulty level is determined by the number of passes in ten samplings of \texttt{DeepSeek-Coder-V2-Lite}. Problems with zero correct solutions are classified as hard, 1-5 correct solutions as medium, and those with more than five as easy.}
\label{method_bench}
\end{table}

\subsection{Automated Workflow}

\begin{figure*}[t]
    \centering
\includegraphics[width=1.0\columnwidth]{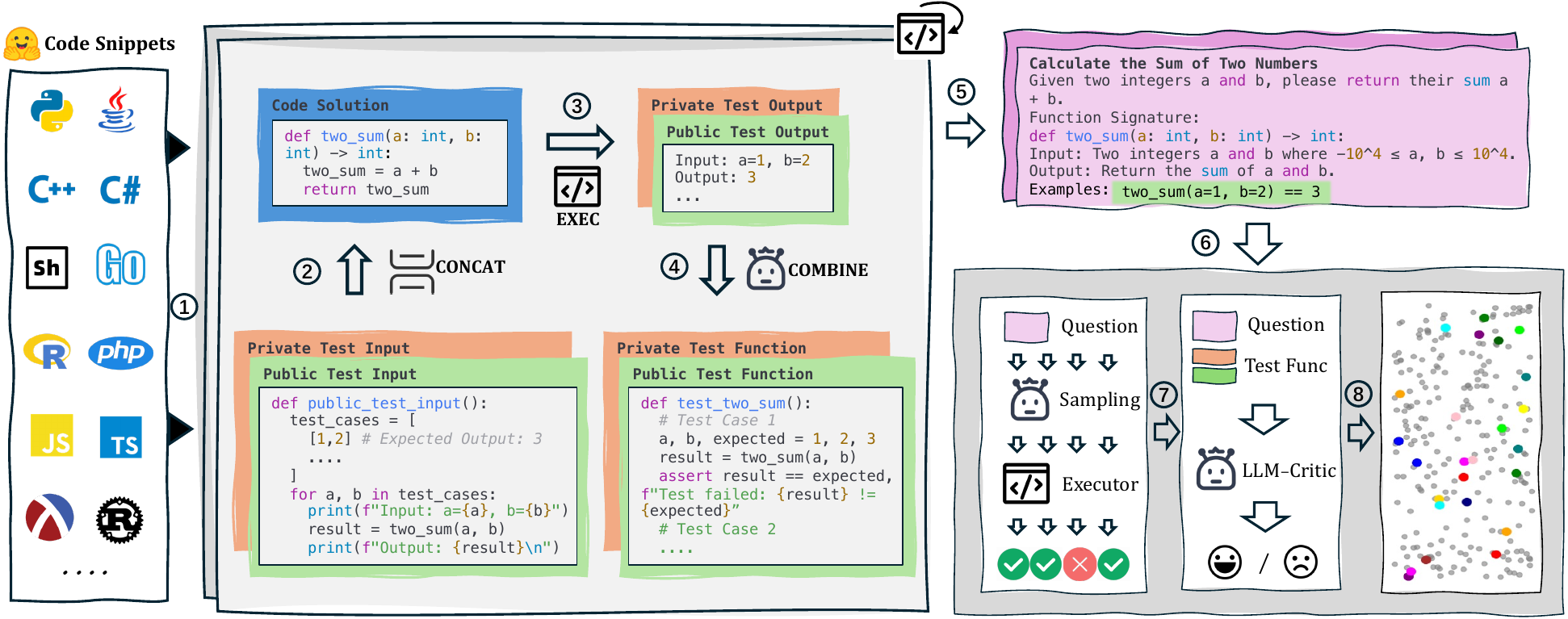}
    \caption{\textbf{The overview of AutoCodeGen}. It first generates code solution and the corresponding public/private test input functions based on multilingual code snippets (\textcircled{1}). They are concatenated and executed in a sandbox to obtain test outputs, which are then combined by the LLM into complete test functions (\textcircled{2},\textcircled{3},\textcircled{4}). Based on the code solution and test function, the LLM is prompted to generate accurate programming problems (\textcircled{5}). Finally, a three-stage data filtering is applied: multiple sampling to remove too easy problems (\textcircled{6}), LLM-as-Critic to discard low-quality ones (\textcircled{7}), and diversity-based tagging to ensure distributional variety (\textcircled{8}).}
	\label{method:workflow}
\end{figure*}

Our AutoCodeGen based on LLM-Sandbox interaction for constructing code generation benchmarks is fully automated. It first generates large-scale multilingual data with guaranteed executability and correctness, then applies a three-stage filtering strategy to ensure the benchmark is challenging, high-quality, and diverse. As illustrated in Figure~\ref{method:workflow}, the workflow includes four key stages: \textbf{Code Solution Generation} (\textcircled{1}), \textbf{Test Function Generation} (\textcircled{2},\textcircled{3},\textcircled{4}), \textbf{Programming Problem Generation} (\textcircled{5}), and \textbf{Data Filtering} (\textcircled{6},\textcircled{7},\textcircled{8}).


\subsubsection{Code Solution Generation}

We begin by extracting multilingual code snippets from Stack-Edu~\citep{allal2025smollm2smolgoesbig}, a large-scale dataset of educational code filtered from The Stack v2~\citep{lozhkov2024starcoder}, as seeds. These seeds span function-level, class-level, and file-level code, sourced from real GitHub repositories, ensuring diversity and practicality. Using a language-specific few-shot prompt, we guide \texttt{DeepSeek-V3-0324} to refine and evolve these seeds into verifiable and self-contained code solutions. During this process, the model removes non-essential logic and adds appropriate comments for clarity. We then validate the correctness of the generated solutions by multilingual sandbox.

\subsubsection{Test Function Generation}



We enhance efficiency and edge-case coverage by first generating test inputs via LLMs and then executing them in a sandbox to obtain the corresponding outputs. Specifically, Test Function Generation is divided into the following three steps:

\textbf{Test Input Generation} The test input functions (both public and private) are generated alongside the above code solution, ensuring alignment between the solution and its inputs. The public test input function includes no more than 3 basic cases and serves demonstration purposes; it will be embedded into the final programming problem as an illustrative usage. In contrast, the private test input function contains 7+ inputs, including edge cases, and functions as the comprehensive test for verifying the correctness of the code solution.

\textbf{Test Output Generation} We concatenate the code solution with test input functions and execute them in the sandbox to obtain the corresponding test outputs.

\textbf{Input-Output Integration} We prompt \texttt{DeepSeek-V3-0324} with both the test input functions and output results to generate coherent and verifiable test functions. Finally, we validate the correctness by executing the code solution together with the generated public and private test functions in the sandbox.

\subsubsection{Programming Problem Generation} 
\label{method_probelm_generate}

Generating high-quality programming problems is challenging, as it requires detailed and accurate problem descriptions. We find that models often omit key information when generating programming problems, such as the entry point specified in the test function. Therefore, we define a set of specifications that generated problems must meet:

\begin{itemize}
    \item Language Specification: Explicitly states the programming language to be used.
    \item Problem Description: A precise and unambiguous description of the task.
    \item Function/Class Naming: Clear identification of all functions and classes involved in the test function.
    \item Input/Output Format: Explicit definitions of input and output types and value ranges.
    \item Example Usage: Provides sample tests embedded with the generated public test functions for reference.
    \item No Solution Hints: The problem description must not include hints to the solution.
\end{itemize}

Using these guidelines, we prompt \texttt{DeepSeek-V3-0324} to generate high-quality programming problems based on the code solution (with appropriate comments) and the corresponding test function, while embedding the public test function as example usage. 

\textbf{Through these three steps, we obtain a large-scale multilingual dataset, where each instance is represented as a tuple \textless{}programming problem, code solution, public test function, private test function\textgreater{}.}

\begin{table}[t]
\small
\centering
\begin{tabular}{cc|cc|cc|cc|cc}
\toprule
\textbf{Origin} & \textbf{Target} & \textbf{Origin} & \textbf{Target} & \textbf{Origin} & \textbf{Target} & \textbf{Origin} & \textbf{Target} & \textbf{Origin} & \textbf{Target} \\ \midrule
 Python          &  R &  Python          &  Ruby  & Java          &  Scala  & Java          &  C\#  &Shell          &  Perl \\
Python          &  Elixir  &Python          &  Julia    & Java         &  Kotlin & JavaScript          &  PHP & C++ & Rust \\ Python          &  Swift   & Python          &  Racket & Java & Dart   &  JavaScript        &  Typescript  \\
\bottomrule
\end{tabular}
\caption{Programming language translation pairs.}
\label{method:translate}
\end{table}

\subsubsection{Data Filtering}
\label{method_data_filtering}

Finally, We apply three filtering and sampling steps to ensure the \textbf{high-difficulty}, \textbf{high-quality}, and \textbf{diversity} of the final benchmark.

\textbf{Difficulty Control} Programming problems that are too simple are not meaningful for evaluating the code generation capabilities of current LLMs. To address this, we employ a moderately capable code model, \texttt{DeepSeek-Coder-V2-Lite}, to filter out too easy problems. Specifically, we sample answers for each problem ten times using the model and validate the correctness via sandbox execution. Problems that are solved correctly in all ten attempts are discarded. Take Python as an example, \texttt{DeepSeek-Coder-V2-Lite} can filter out 25.1\% of overly simple problems.

\textbf{Quality Control} During the aforementioned problem generation stage, we define six specifications to guide the generation of detailed and accurate programming problems. To further ensure high quality, we employ \texttt{DeepSeek-R1-0528} to critique each \textless{}problem, test function\textgreater{} pair. Specifically, assuming the problem is entirely correct, we evaluate the test function’s consistency with problem based on the following seven concepts:
\begin{itemize}
    \item Whether the function/class names or signatures match the problem description.
    \item Whether the test cases involve randomness or are non-reproducible.
    \item Whether the testing objective aligns with the stated purpose of the problem.
    \item Whether numerical precision is handled appropriately.
    \item Whether exception handling is used in a way that invalidates the test cases.
    \item Whether the test cases check for requirements beyond the problem description.
    \item Whether the test cases are comprehensive.
\end{itemize}

\textbf{Diversity Sampling}
We aim for our benchmark to cover as many real-world scenarios as possible. To this end, we perform diversity-based sampling on the existing data to construct the final benchmark. We use \texttt{DeepSeek-V3-0324} to label each problem. We then divide the problems into different pools by category and perform cyclic sampling, ensuring a broad representation of programming scenarios.

\subsubsection{Approximate Language Translation}

For Python, C++, Shell, Java, JavaScript, and Go, we directly use the workflow described above. For the other 14 languages, while the proposed workflow is still applicable, we choose to employ an approximate language translation approach due to their limited data resources and lack of diversity. We extract unused data from the data pool generated in Section~\ref{method_probelm_generate} and translate them into the target low-resource language, as shown in Table~\ref{method:translate}. This ensures a sufficient and diverse dataset, which is further refined through the Data Filtering process, as described in Section~\ref{method_data_filtering}.

\subsubsection{AutoCodeBench-Lite Construction}

To facilitate quicker and more efficient model evaluations, we create the AutoCodeBench-Lite, a simplified subset of AutoCodeBench.
Specifically, we collect the problem-solving results from all models and sort the problems in ascending order based on the number of passes. After discarding problems with fewer than 2 passes, we select approximately 1,500 problems based on their pass count in ascending order. These problems, which have been solved correctly by existing models at least twice and have a certain level of difficulty, are selected to amplify the differences between the models. We use these problems as the set for the Lite version.

\section{Experiments}

\begin{table}[]
\resizebox{\textwidth}{!}{
\begin{tabular}{lccccccccccccccccccccc}
\toprule 
& \cellcolor[HTML]{D9D9D9}\textbf{Average}                              & \textbf{Python}                               & \textbf{Cpp}                                  & \textbf{Java}                                 & \textbf{JS}                                   & \textbf{Go}                                   & \textbf{Shell}                                & \textbf{Csharp}                               & \textbf{Dart}                                 & \textbf{Elixir}                               & \textbf{Julia}                                & \textbf{Kotlin}                               & \textbf{Perl}                                 & \textbf{PHP}                                  & \textbf{Racket}                               & \textbf{R}                                    & \textbf{Ruby}                                 & \textbf{Rust}                                 & \textbf{Scala}                                & \textbf{Swift}                                & \textbf{TS}                                   \\
\bottomrule
\end{tabular}
}
\caption{Pass@1 (\%) performance of different models for AutoCodeBench. \textbf{\textit{\textcolor{red}{Current Upper Bound}}} represents the Pass@1 value calculated by taking the union of problems correctly solved by all models.}
\label{exp:acb}
\end{table}

\begin{table}[]
\resizebox{\textwidth}{!}{
\begin{tabular}{lccccccccccccccccccccc}
\toprule
\textbf{}                                                      & \cellcolor[HTML]{D9D9D9}\textbf{Average}                               & \textbf{Python}                                & \textbf{Cpp}                                   & \textbf{Java}                                  & \textbf{JS}                                    & \textbf{Go}                                    & \textbf{Shell}                                 & \textbf{Csharp}                                & \textbf{Dart}                                  & \textbf{Elixir}                                & \textbf{Julia}                                 & \textbf{Kotlin}                                & \textbf{Perl}                                  & \textbf{PHP}                                   & \textbf{Racket}                                & \textbf{R}                                     & \textbf{Ruby}                                  & \textbf{Rust}                                  & \textbf{Scala}                                 & \textbf{Swift}                                 & \textbf{TS}                                    \\
\bottomrule
\end{tabular}
}
\caption{Pass@1 (\%) performance of different models for AutoCodeBench-Lite.}
\label{exp:acb_lite}
\end{table}

\subsection{Experimental Setup}

\textbf{LLMs} We evaluate a diverse set of open-source models with sizes ranging from 1.5B to 1T parameters, as well as leading proprietary models on AutoCodeBench. These models are classified based on their families:

\begin{itemize}
    \item \textbf{OpenAI}: \texttt{o3} and \texttt{o4-mini}~\citep{o3}; \texttt{GPT4.1}~\citep{GPT-4.1}, and \texttt{GPT4o}~\citep{gpt4o}.

    \item \textbf{Claude}: \texttt{Claude Opus 4} and \texttt{Claude Sonnet 4}~\citep{claude4}.

    \item \textbf{Gemini}: \texttt{Gemini 2.5 Pro} and \texttt{Gemini 2.5 Flash}~\citep{gemini2.5}.

    \item \textbf{DeepSeek}: \texttt{DeepSeek-R1-0528}~\citep{deepseekai2025deepseekr1}, \texttt{DeepSeek-V3-0324}~\citep{deepseek_v3} and \texttt{DeepSeek-Coder} Series~\citep{deepseekai2024deepseekcoderv2,guo2024deepseekcoder}.

    \item \textbf{Hunyuan}: \texttt{Hunyuan-TurboS} and \texttt{Hunyuan-Coder-7B-Preview}~\citep{tencenthunyuanteam2025hunyuanturbos}.

    \item \textbf{Qwen}: \texttt{Qwen3-235B-A22B-Thinking-2507}, \texttt{Qwen3-235B-A22B-Instruct-2507}, and \texttt{Qwen3} Series~\citep{yang2025qwen3}; \texttt{Qwen3-Coder-480B-A35B-Instruct}~\citep{qwen3coder}, \texttt{Qwen2.5-Coder} Series~\citep{hui2024qwen25codertechnicalreport} and \texttt{Qwen2.5-72B}~\citep{qwen2025qwen25technicalreport}

    \item \textbf{Seed}: \texttt{Seed1.6-Thinking}~\citep{seed1.6}, \texttt{Seed1.6}~\citep{seed1.6} and \texttt{Seed-Coder-8B}~\citep{seed2025seedcoderletcodemodel}.

    \item \textbf{GLM}: \texttt{GLM-4.5} and \texttt{GLM-4.5-Air}~\citep{glm4p5}.

    \item \textbf{Other Models}: \texttt{ERNIE-X1-Turbo-32K}~\citep{erniex1}, \texttt{Kimi-K2}~\citep{kimiteam2025kimik2openagentic}, and \texttt{OpenCoder-}\texttt{8B} \citep{huang2025opencoderopencookbooktoptier}.
\end{itemize}


\textbf{Evaluation Details} We use the Pass@1~\citep{chen2021evaluatinglargelanguagemodels} as the default evaluation metric. In terms of inference parameters, for proprietary LLMs, \texttt{Qwen3-235B-A22B-Thinking-}\texttt{2507}, \texttt{Qwen3-235B-A22B-Instruct-2507}, \texttt{Kimi-K2}, \texttt{Qwen3-Coder-480B-A35B-Instruct}, and \texttt{Hunyuan-TurboS}, we directly call their APIs without any additional parameters. For \texttt{DeepSeek-V3-0324}, \texttt{DeepSeek-R1-0528}, \texttt{Seed-Coder-8B}, the \texttt{Qwen3} series, we use their officially recommended parameters. The remaining models use greedy decoding with temperature set to 0. All non-APIs models are deployed using the \texttt{VLLM}~\citep{kwon2023efficient} framework. All models are provided with our custom system prompt, which standardizes the output format: \textbf{\textit{You are an expert programmer. Your task is to provide a code solution within a single Markdown code block for the given programming problem. Do not include any direct execution commands, test cases, or usage examples within the code block.}}

\subsection{Main Results}

\begin{figure*}[t]
    \centering
\includegraphics[width=1.0\columnwidth]{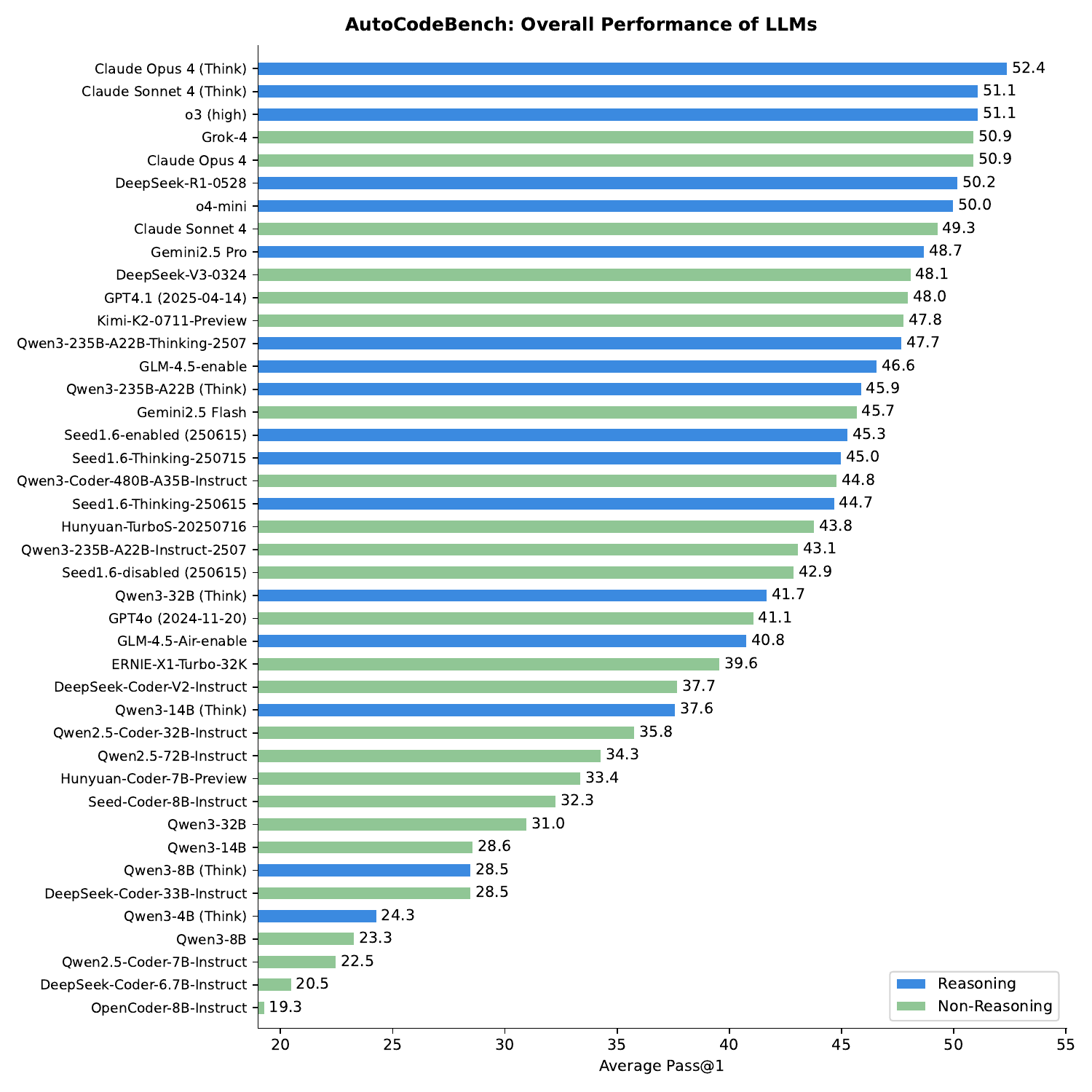}
    \caption{AutoCodeBench leaderboard showing Pass@1 performance of various LLMs.}
	\label{exper:leaderboard}
\end{figure*}

We comprehensively evaluate the performance on AutoCodeBench (ACB) and AutoCodeBench-Lite (ACB-Lite), with results across different programming languages shown in Tables~\ref{exp:acb} and~\ref{exp:acb_lite}. The corresponding leaderboards are shown in Figures~\ref{exper:leaderboard} and~\ref{appendix:leaderboard_lite}.

\textbf{ACB Exhibits High Difficulty.} None of the evaluated models achieve an average score above 53, highlighting the high difficulty of the tasks in ACB. This underscores that the benchmark is specifically designed to challenge current models, pushing the limits of their code generation capabilities. The results further indicate that these models still struggle to solve complex, practical multilingual problems effectively.

\textbf{ACB-Lite amplifies the performance differences between models.} Since ACB-Lite is derived by filtering problems based on the number of passes from all models, problems solved by the majority of models are excluded. Therefore, ACB-Lite amplifies the performance differences between models compared to ACB, making it more effective for comparing model performance.

\textbf{Claude Opus 4 Shows State-of-the-art Performance.} Regardless of whether the mode are reasoning or non-reasoning, \texttt{Claude Opus 4} consistently ranks first in ACB(-Lite). This highlights the superior performance of Claude models in addressing diverse and complex tasks that involve practical coding scenarios across multiple languages. This conclusion aligns with the results observed in SWE-bench~\citep{jimenez2024swebench} and Multi-SWE-bench~\citep{zan2025multiswebenchmultilingualbenchmarkissue}.

\textbf{The Reasoning Mode Helps LLMs Solve Multilingual Challenges.} Overall, reasoning-based models perform better than non-reasoning models across various programming languages. We believe that reasoning and thinking in the reasoning mode help to solve the complex, multi-logical problems in ACB.


\textbf{Upper Bound Reveals Models' Complementary Strengths and Improvement Potential.} Although all models exhibit moderate performance (below 53) on the ACB benchmark, their combined upper bound reaches 74.8, highlighting the potential for models to learn from each other. The state-of-the-art model only achieves 64.5 pass@1 on ACB-Lite further emphasizes this, as it suggests that no single model excels across all problems. Additionally, the fact that no model dominates across all languages underscores the varying multilingual capabilities of each model, revealing significant room for improvement.

\subsection{Performance Across Popular and Low-Resource Programming Languages}


\begin{figure}[t]
\centering
\includegraphics[width=0.7\textwidth]{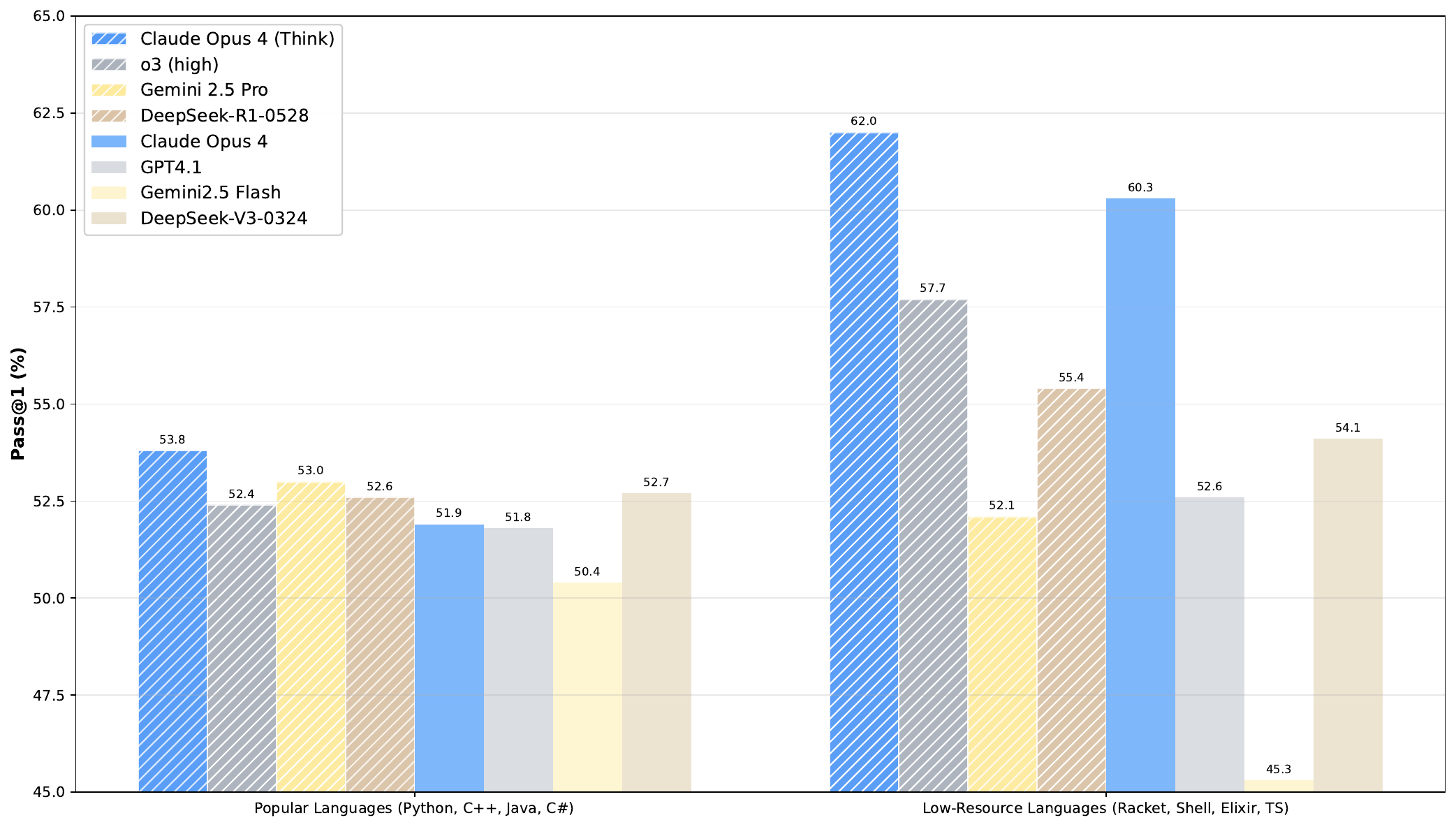}
\caption{The performance comparison of different models across two language sets.}
\label{exp:compare_popular_longtail}
\end{figure}



We select four popular languages (Python, C++, Java, C\#) and four low-resource languages (Racket, Shell, Elixir, TS) based on the TIOBE Index~\footnote{\url{https://tiobe.org.cn/tiobe-index/}}to evaluate model performance across different language scenarios. As shown in Figure~\ref{exp:compare_popular_longtail}, the difference in average Pass@1 scores among the models for popular languages is small, with scores ranging from 50.4 to 53.8. This indicates that they have been adequately trained on these widely-used languages. However, when faced with low-resource languages, the performance gap between models from different families widens (ranging from 45.3 to 62.0). \texttt{Claude Opus 4} outperforms other models significantly in both reasoning and non-reasoning modes. This result highlights the insufficient attention given to low-resource programming languages in the development of most models.

Besides, since we use the moderately capable \texttt{DeepSeek-Coder-V2-Lite} to filter simple problems, the Pass@1 scores of top models on popular languages are relatively low. However, these models perform significantly better on low-resource languages. This indicates that the performance gap between models of different sizes is more pronounced on low-resource languages, likely because \texttt{DeepSeek-Coder-V2-Lite} struggles to filter out simple problems in these scenarios due to its limited capability in handling low-resource languages.

\subsection{Performance Across Multi-Logic Programming Problems}

\begin{figure*}[t]
    \centering
\includegraphics[width=0.7\columnwidth]{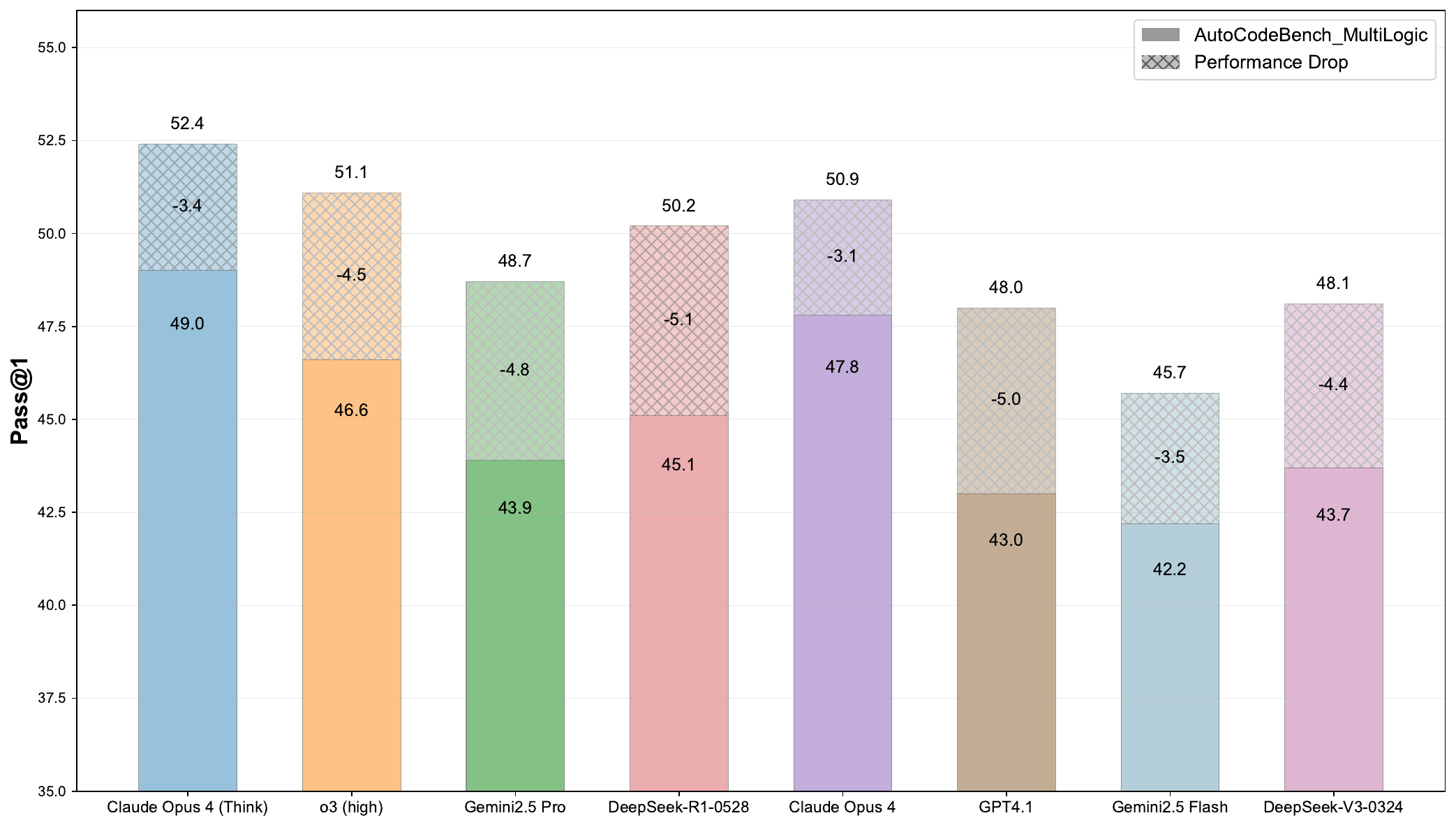}
    \caption{Performance drop of models on multi-logic problems (1,622) compared to full dataset.}
	\label{exp:multilogic}
\end{figure*}

One key feature of AutoCodeBench, compared to previous benchmarks, is the inclusion of multi-logical problems. These problems require models to implement multiple distinct functions or classes within a single task, challenging their ability to handle multiple core demands simultaneously. 

We use \texttt{DeepSeek-V3-0324} to identify all multi-logical problems in AutoCodeBench and evaluate model performance on them. The results, shown in Figure~\ref{exp:multilogic}, reveal a significant performance drop for all models when faced with multi-logical tasks. \texttt{Claude Opus 4} shows a relatively smaller decline, while the \texttt{DeepSeek} series, \texttt{Gemini2.5 Pro}, and \texttt{GPT 4.1} exhibit larger drops. This indicates that current models need to enhance their ability to process multi-logical problems, which is particularly crucial in real-world code agent scenarios.

\subsection{Performance Analysis of Scaling Laws}


\begin{figure}[t]
\centering
\begin{subfigure}{0.48\textwidth}
    \centering
    \includegraphics[width=\textwidth]{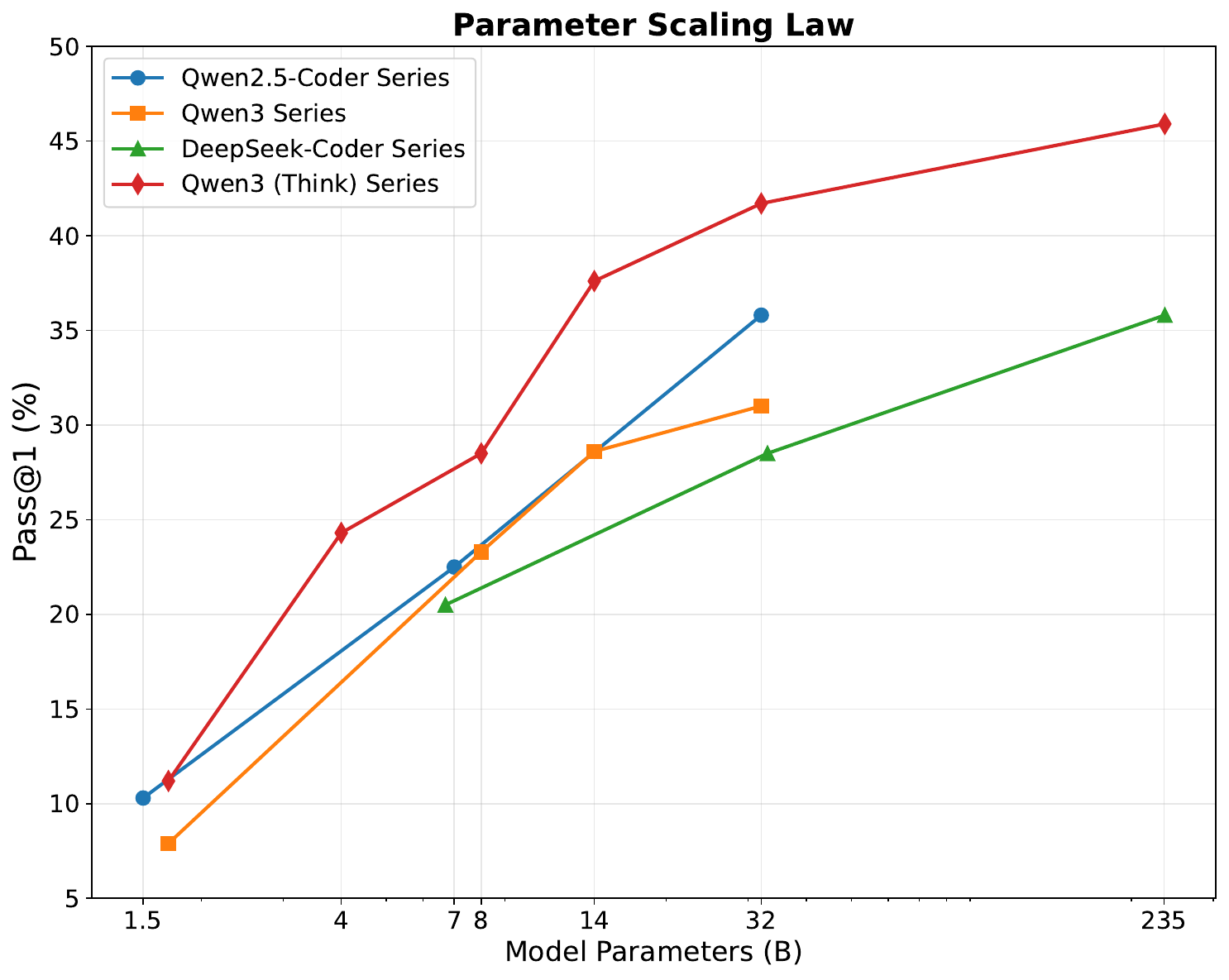}
    \label{fig:scaling}
\end{subfigure}
\hfill
\begin{subfigure}{0.48\textwidth}
    \centering
    \includegraphics[width=\textwidth]{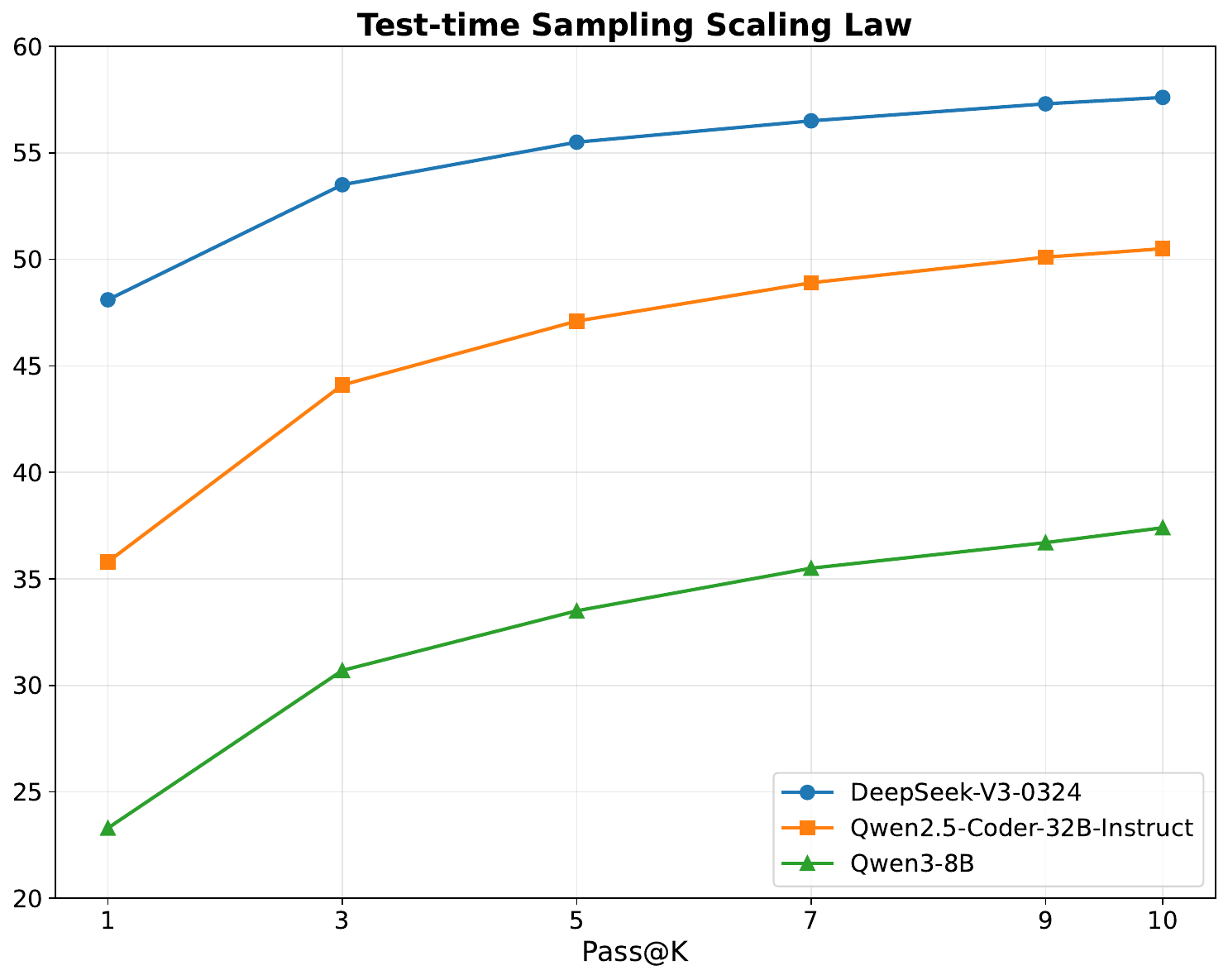}
    \label{fig:sampling_scaling}
\end{subfigure}
\caption{Scaling laws for different models.}
\label{exp:scaling}
\end{figure}


Figure~\ref{exp:scaling} compares parameter scaling and test-time sampling scaling across different models. The parameter scaling law (left) shows significant variation between models, with \texttt{Qwen3 (Think)} Series demonstrating the steepest scaling curve, indicating that chain-of-thought reasoning particularly benefits larger models. The test-time sampling scaling law (right) reveals more uniform behavior, with three models showing similar improvement rates from increased sampling during inference. These results suggest that while test-time sampling provides consistent benefits regardless of model size, reasoning capabilities scale more aggressively with model size.

\subsection{Performance Analysis of Multi-Turn Refinement with Sandbox Feedback}

\begin{figure}[t]
\centering
\includegraphics[width=0.7\textwidth]{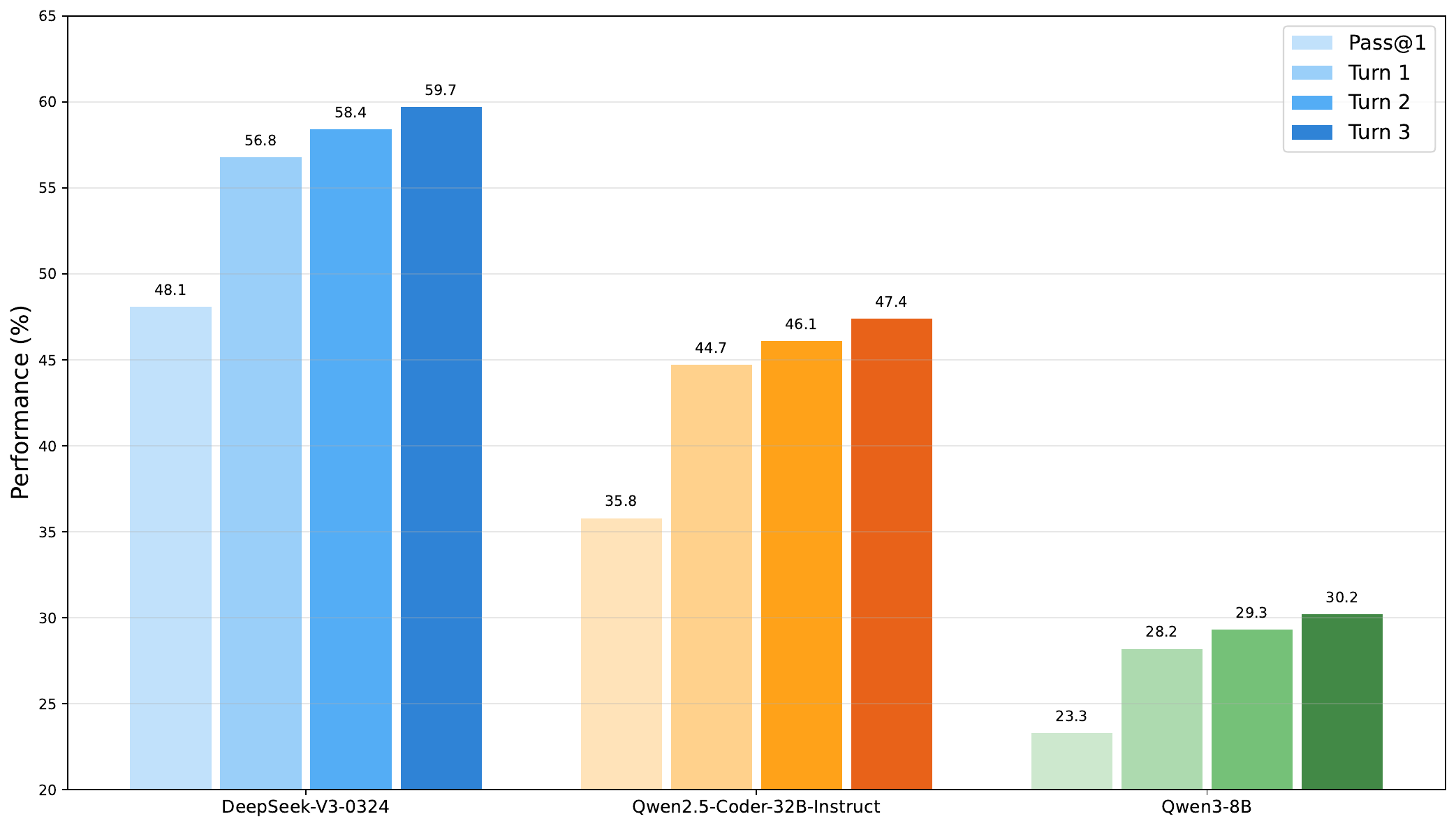}
\caption{Performance improvement across multi-turn refinement with sandbox feedback.}
\label{exp:feedback}
\end{figure}

As shown in Figure~\ref{exp:feedback}, we evaluate how models leverage execution error messages to iteratively refine their code solutions. The results highlight the substantial value of our multilingual sandbox error feedback across all evaluated models. \texttt{DeepSeek-V3-0324} achieves remarkable improvement from 48.1\% to 59.7\% after three refinement turns. \texttt{Qwen2.5-Coder-32B-Instruct} demonstrates even greater relative gains, advancing from 35.8\% to 47.4\%, while \texttt{Qwen3-8B} shows consistent progress from 23.3\% to 30.2\%. The most significant performance gains occur during the first refinement turn, with diminishing returns in subsequent iterations. This pattern suggests that models can effectively leverage execution feedback to identify and correct common coding errors, though the complexity of remaining problems increases with each iteration. The consistent improvement across different model scales indicates that multi-turn refinement with sandbox feedback is a valid strategy for enhancing code generation quality.

\subsection{AutoCodeBench-Complete: Evaluating Base Model Capabilities}

\begin{table}[]
\resizebox{\textwidth}{!}{
\begin{tabular}{lccccccccccccccccccccc}
\toprule
                         & \cellcolor[HTML]{D9D9D9}\textbf{Average} & \textbf{Python}           & \textbf{Cpp}              & \textbf{Java}             & \textbf{JS}               & \textbf{Go}               & \textbf{Shell}            & \textbf{Csharp}           & \textbf{Dart}             & \textbf{Elixir}           & \textbf{Julia}            & \textbf{Kotlin}           & \textbf{Perl}             & \textbf{PHP}              & \textbf{Racket}           & \textbf{R}                & \textbf{Ruby}             & \textbf{Rust}             & \textbf{Scala}            & \textbf{Swift}            & \textbf{TS}               \\
\textbf{Count}           & \textbf{}                                & {\color[HTML]{7F7F7F} 50} & {\color[HTML]{7F7F7F} 50} & {\color[HTML]{7F7F7F} 50} & {\color[HTML]{7F7F7F} 50} & {\color[HTML]{7F7F7F} 50} & {\color[HTML]{7F7F7F} 50} & {\color[HTML]{7F7F7F} 50} & {\color[HTML]{7F7F7F} 50} & {\color[HTML]{7F7F7F} 50} & {\color[HTML]{7F7F7F} 50} & {\color[HTML]{7F7F7F} 50} & {\color[HTML]{7F7F7F} 50} & {\color[HTML]{7F7F7F} 50} & {\color[HTML]{7F7F7F} 50} & {\color[HTML]{7F7F7F} 50} & {\color[HTML]{7F7F7F} 50} & {\color[HTML]{7F7F7F} 50} & {\color[HTML]{7F7F7F} 50} & {\color[HTML]{7F7F7F} 50} & {\color[HTML]{7F7F7F} 50} \\
\midrule
\multicolumn{22}{c}{\textbf{30B+ Models}}                                                                                                                                                                                                                                                                                                                                                                                                                                                                                                                                                                                                           \\
DeepSeek-Coder-V2-Base   & \cellcolor[HTML]{D9D9D9}\textbf{39.0}    & 24.0                      & 32.0                      & 40.0                      & 44.0                      & 34.0                      & 26.0                      & 64.0                      & 38.0                      & 52.0                      & 46.0                      & 56.0                      & 38.0                      & 36.0                      & 32.0                      & 26.0                      & 40.0                      & 26.0                      & 36.0                      & 42.0                      & 48.0                      \\
Qwen2.5-Coder-32B        & \cellcolor[HTML]{D9D9D9}35.5             & 36.0                      & 34.0                      & 32.0                      & 32.0                      & 38.0                      & 34.0                      & 58.0                      & 30.0                      & 42.0                      & 38.0                      & 52.0                      & 40.0                      & 32.0                      & 30.0                      & 26.0                      & 30.0                      & 18.0                      & 30.0                      & 34.0                      & 44.0                      \\
Qwen2.5-72B              & \cellcolor[HTML]{D9D9D9}35.9             & 32.0                      & 22.0                      & 38.0                      & 40.0                      & 22.0                      & 34.0                      & 62.0                      & 22.0                      & 42.0                      & 38.0                      & 46.0                      & 42.0                      & 46.0                      & 28.0                      & 26.0                      & 38.0                      & 28.0                      & 28.0                      & 30.0                      & 54.0                      \\
\midrule
\multicolumn{22}{c}{\textbf{$\sim$8B Models}}                                                                                                                                                                                                                                                                                                                                                                                                                                                                                                                                                                                                       \\
Seed-Coder-8B-Base       & \cellcolor[HTML]{D9D9D9}\textbf{31.6}    & 26.0                      & 22.0                      & 40.0                      & 30.0                      & 32.0                      & 12.0                      & 54.0                      & 24.0                      & 48.0                      & 30.0                      & 48.0                      & 28.0                      & 36.0                      & 22.0                      & 26.0                      & 32.0                      & 18.0                      & 20.0                      & 36.0                      & 48.0                      \\
OpenCoder-8B-Base        & \cellcolor[HTML]{D9D9D9}26.1             & 22.0                      & 6.0                       & 28.0                      & 34.0                      & 30.0                      & 24.0                      & 52.0                      & 10.0                      & 42.0                      & 32.0                      & 28.0                      & 26.0                      & 24.0                      & 20.0                      & 20.0                      & 28.0                      & 14.0                      & 26.0                      & 14.0                      & 42.0                      \\
Qwen2.5-Coder-7B         & \cellcolor[HTML]{D9D9D9}24.6             & 20.0                      & 10.0                      & 22.0                      & 28.0                      & 24.0                      & 14.0                      & 46.0                      & 8.0                       & 46.0                      & 32.0                      & 42.0                      & 30.0                      & 30.0                      & 14.0                      & 20.0                      & 18.0                      & 16.0                      & 24.0                      & 14.0                      & 34.0                      \\
DeepSeek-Coder-6.7B-Base & \cellcolor[HTML]{D9D9D9}22.9             & 20.0                      & 14.0                      & 26.0                      & 34.0                      & 18.0                      & 18.0                      & 50.0                      & 8.0                       & 44.0                      & 20.0                      & 38.0                      & 28.0                      & 18.0                      & 12.0                      & 14.0                      & 34.0                      & 6.0                       & 10.0                      & 4.0                       & 42.0                      \\
Qwen3-8B-Base            & \cellcolor[HTML]{D9D9D9}22.6             & 20.0                      & 14.0                      & 18.0                      & 34.0                      & 20.0                      & 12.0                      & 50.0                      & 6.0                       & 34.0                      & 26.0                      & 24.0                      & 32.0                      & 30.0                      & 8.0                       & 20.0                      & 30.0                      & 8.0                       & 14.0                      & 16.0                      & 36.0                     \\
\bottomrule
\end{tabular}
}
\caption{Pass@1 (\%) performance of different base models for 3-shot AutoCodeBench-Complete.}
\label{exp:acb_comp}
\end{table}

Existing benchmarks for evaluating base models on code generation tasks, such as HumanEval~\citep{chen2021evaluatinglargelanguagemodels}, MBPP~\citep{austin2021programsynthesislargelanguage}, and their multilingual counterpart MultiPL-E~\citep{cassano2022multiplescalableextensibleapproach}, primarily emphasize simple algorithmic problems. They fail to comprehensively measure the programming capabilities of base models in diverse, real-world scenarios. Building on the diversity and comprehensiveness of AutoCodeBench, we present \textbf{AutoCodeBench-Complete}, a completion-based evaluation benchmark tailored to assess the code generation capabilities of base models. To build this benchmark, we select 1,000 datas from ACB-Lite, ensuring a balanced distribution of 50 problems per programming language. We use 3-shot demonstrations to evaluate the performance of base models.

Table~\ref{exp:acb_comp} presents the performance of various base models on ACB-Complete. Among models with 8B parameters or fewer, \texttt{Seed-Coder-8B-Base} demonstrates superior performance, consistent with its strong showing on ACB(-Lite). This consistency suggests that the pretraining process effectively equipped \texttt{Seed-Coder} models with strong multilingual programming capabilities, enabling them to handle diverse coding scenarios across multiple languages. Additionally, an interesting observation arises when comparing \texttt{Qwen2.5-Coder-7B} and \texttt{OpenCoder-8B}. While \texttt{Qwen2.5-Coder-}\texttt{7B-Instruct} outperforms \texttt{OpenCoder-8B-Instruct} on ACB(-Lite), the trend reverses on ACB-Complete. We believe that \texttt{Qwen2.5-Coder-} \texttt{7B-Instruct} incorporate a more comprehensive multilingual code dataset during its post-training phase, thereby enhancing its code understanding and generation abilities.

We hope that AutoCodeBench-Complete will serve as a comprehensive benchmark for evaluating the code generation capabilities of base models. We believe it will provide researchers and practitioners with a testbed that more closely mirrors the complexity and diversity of real-world programming tasks.

\section{Further Discussion}

\subsection{Manual Verification}

In our automated workflow, we employ carefully designed problem specifications and an LLM-as-Critic mechanism to enforce automated quality control. However, since LLMs cannot guarantee 100\% accuracy, the overall quality of AutoCodeBench remains uncertain. To address this, we employ 6 professional annotators to quantitatively assess the quality of AutoCodeBench. Specifically, we develop a visual annotation interface that displays each data instance, including the programming question, its corresponding test function, and the annotation contents from \texttt{DeepSeek-R1-0528}. Based on this interface, annotators are asked to assess the correctness of the test function and its alignment with the problem description. They then assign a binary label (yes/no) indicating whether the instance is valid. We select six programming languages for this verification: Python, C++, Java, JavaScript, Go, and Shell. The results show that AutoCodeBench achieves a \textbf{87.6\%} accuracy rate, demonstrating the reliability and feasibility of our automated benchmark construction process. In comparison, the model upper bounds and the performance of Claude Opus 4 (Reasoning) in these six languages are only 66.9(\textcolor{red}{$\Delta20.7$}) and 44.6(\textcolor{red}{$\Delta43.0$}), respectively, highlighting the significant potential for improvement. More detailed analysis is provided in Appendix~\ref{appendix:b}.

\subsection{Hypotheses on Model Bias in the Generation Process}

\begin{table}[]
\resizebox{\textwidth}{!}{
\begin{tabular}{lcccccc}
\toprule
    & \textbf{Initial Stage} (Rank) & \textbf{After Simple Problem Filtering} (Rank) & \textbf{After Critic Filtering} (Rank) \\
\midrule
DeepSeek-V3-0324  & 47.1 (3)    & 25.7 $_{-21.4}$ (4)                     & 31.6 $_{+5.9}$ (4)    \\
DeepSeek-R1-0528  & 48.9   (2)    & 28.7 $_{-20.2}$                            (2)    & 36.2 $_{+7.5}$                     (2)    \\
o3                & 46.4                   (4)    & 28.1 $_{-18.3}$                            (3)    & 34.9 $_{+6.8}$                     (3)    \\
Gemini2.5 Pro     & 51.4                  (1)    & 31.6 $_{-19.8}$                            (1)    & 38.7 $_{+7.0}$                     (1)    \\
Qwen2.5-Coder-32B-Instruct & 39.9                  (5)    & 17.1 $_{-22.8}$                          (5)    & 22.0 $_{+4.9}$                       (5)   \\
\bottomrule
\end{tabular}
}
\caption{The average pass@1 scores and rankings of models at different stages.}
\label{exp:bias}
\end{table}

It is well-known that models exhibit inherent biases, particularly their tendency to favor their own outputs—a common phenomenon in automated data synthesis and evaluation tasks~\citep{panickssery2024llm,chen2025llmevaluatorspreferreason}. Our automated workflow is no exception to this issue. While completely eliminating such bias is challenging, we employ several mitigation strategies. Specifically, we intentionally only use DeepSeek series models in the workflow to prevent bias from affecting other model families. We hypothesize that using \texttt{DeepSeek-V3-0324} for code generation and \texttt{DeepSeek-R1-0528} for the Critic process may introduce favorable bias toward DeepSeek families. To counteract this, we employ \texttt{DeepSeek-Coder-V2-Lite} during the simple problem filtering phase, creating a "push-and-pull" mechanism that balances potential biases across different stages.

To quantitatively assess bias, we sampled 3,600 data points across six programming languages (Python, C++, Java, JS, Go, and Shell) and tracked performance changes at each generation stage, as shown in Table~\ref{exp:bias}. The results reveal nuanced bias patterns: simple problem filtering negatively impacts smaller models (\texttt{Qwen2.5-Coder-32B-Instruct}) more than DeepSeek series, while the Critic process benefits \texttt{DeepSeek-R1-0528} but surprisingly provides greater improvements to reasoning models (\texttt{o3} and \texttt{Gemini 2.5 Pro}) than to \texttt{DeepSeek-V3-0324}. This suggests that model bias depends not only on model family but also on factors like model size and reasoning modes.  Furthermore, as mutual distillation between models from different families continues, this bias becomes increasingly difficult to measure. In conclusion, we believe that our automated process may introduce a favorable bias toward the DeepSeek family of models, but the impact is minimal.
\section{Related Work}

\subsection{Code Generation Benchmarks}
The rapid evolution of Code Large Language Models (LLMs)—spanning open-source models like CodeLLama~\citep{roziere2023code}, DeepSeek-Coder~\citep{zhu2024deepseek}, and Qwen-Coder~\citep{hui2024qwen2}, as well as proprietary systems like Claude~\citep{claude4} series, GPT~\citep{gpt4o,GPT-4.1} series, and the Gemini~\citep{gemini2.5} family—has fundamentally reshaped software development. This progress necessitates robust and contemporary benchmarks to accurately assess their capabilities in tasks such as code generation and debugging. Pioneering benchmarks like HumanEval~\citep{chen2021evaluatinglargelanguagemodels} and MBPP~\citep{austin2021programsynthesislargelanguage} established a foundation by evaluating functional correctness on short, algorithm-centric Python problems. However, they are hampered by limitations such as potential data contamination, narrow programming language coverage, and a disconnect from real-world applications. To overcome these shortcomings, subsequent benchmarks~\citep{jimenez2024swebench,jain2025livecodebench,wang2025ojbench,guo2025codeeditorbenchevaluatingcodeediting,zhang2025codecriticbenchholisticcodecritique,zhang2025artifactsbench} have targeted more complex programming tasks. A notable line of work focuses on algorithmic challenges from programming contests~\citep{hendrycks2021measuring,Li_2022,jain2025livecodebench,wang2025ojbench,zheng2025livecodebenchproolympiadmedalists}. LiveCodeBench~\citep{jain2025livecodebench} mitigates data contamination by continuously sourcing new problems from competitive programming platforms. OJBench~\citep{wang2025ojbench} presents a rigorous, competition-level benchmark with 232 problems from prestigious contests like NOI and ICPC, demanding advanced code reasoning. APPS~\citep{hendrycks2021measuring} offers a large-scale dataset of 10,000 problems categorized by difficulty. Another stream of research evaluates more comprehensive and multilingual capabilities~\citep{cassano2022multiplescalableextensibleapproach,peng2024humanevalxlmultilingualcodegeneration,zhang2024naturalcodebenchexaminingcodingperformance,chai2025mceval,fullstackbench}. McEval~\citep{chai2025mceval} is a massively multilingual benchmark covering 40 languages for generation, explanation, and completion tasks. FullStackBench~\citep{fullstackbench} assesses LLMs in realistic, multi-domain scenarios across 16 languages, employing a novel execution environment. Besides, some works propose more sophisticated evaluation frameworks. For example, ArtifactsBench~\citep{zhang2025artifactsbench} introduces automated multimodal evaluation for visual code generation, while CodeCriticBench~\citep{zhang2025codecriticbenchholisticcodecritique} focuses on holistic code critique evaluation. A common thread across these benchmarks is their reliance on labor-intensive manual curation for collecting problems, authoring ground-truth solutions, and designing test cases.

In contrast, our proposed work introduces a fully automated, dynamically constructed evaluation system. Unlike traditional benchmarks, it eliminates manual intervention, enabling a more scalable, consistent, and continuously evolving evaluation. Furthermore, it is designed to maintain a balanced difficulty distribution and support multiple languages, addressing the increasing demand for evaluating LLMs in diverse programming languages and contexts.

\subsection{Code Data Synthesis}
To reduce dependence on manually curated data, a growing body of research explores automatic data synthesis to augment the training of Code LLMs~\citep{luo2024wizardcoder,magicoder,zheng-etal-2024-opencodeinterpreter,wu2024inversecoderselfimprovinginstructiontunedcode,yu2024wavecoderwidespreadversatileenhancement,ahmad2025opencodeinstructlargescaleinstructiontuning,xu2025kodcode}. For instance, Evol-Instruct~\citep{luo2024wizardcoder} uses heuristic prompts to guide LLMs in evolving existing programming problems, thereby increasing their diversity and difficulty. OSS-Instruct~\citep{magicoder} prompts LLMs to generate new coding problems and solutions from raw, open-source code snippets. KodCode~\citep{xu2025kodcode} synthesizes a broad spectrum of Python coding tasks—including questions, solutions, and test cases—and ensures correctness through a systematic self-verification procedure. Some other methods focus on model self-improvement~\citep{wu2024inversecoder,NEURIPS2024selfcodealign,chen-etal-2025-revisit,zhou2025refinecoderiterativeimprovinglarge}. Inverse-Instruct~\citep{wu2024inversecoder} is a self-improvement technique that generates new instructions by "back-translating" code from an LLM's own training set, reducing the need to distill from more powerful proprietary models. Similarly, SelfCodeAlign~\citep{NEURIPS2024selfcodealign} introduces a pipeline for self-aligning code LLMs without extensive human annotation, using the same base model for both data generation and validation. Collectively, these data synthesis methods significantly reduce the reliance on manual curation and enable the continuous expansion of the problem space for training. Our work extends this paradigm of automation from data augmentation to the benchmark creation process. By leveraging extensive LLM-sandbox interaction, our pipeline not only automates the synthesis of verifiable test problems but can also be naturally repurposed for synthesizing high-quality training datasets.

\section{Conclusion}
\label{sec:conclusion}

In this paper, we introduce AutoCodeGen, an automated workflow based on LLM-Sandbox interaction, designed to generate multilingual verifiable code data without any manual annotation. Through this novel approach, we have successfully built AutoCodeBench, a large-scale, human-free code generation benchmark. AutoCodeBench contains 3,920 problems, evenly distributed across 20 programming languages, and is characterized by its high difficulty, practicality, and diversity. We also provide AutoCodeBench-Lite (a simplified version) and AutoCodeBench-Comp (a completion-based benchmark specifically designed for base models). Furthermore, we open-sourced a multilingual sandbox that supports over 20+ programming languages to enable high-concurrency evaluation and training. Our evaluation of more than 30 mainstream open-source and proprietary LLMs reveals that even the most advanced models still face challenges when confronted with the complex and diverse multilingual tasks set by AutoCodeBench, especially when handling multi-logic scenarios. We hope that the AutoCodeBench series will become a valuable resource, inspiring the community to focus on more challenging and practical multilingual code generation scenarios. In addition, our comprehensive analysis of AutoCodeGen and AutoCodeBench provides valuable insights for the future development of code generation benchmarks.


\section{Acknowledgements}

In addition to all the authors of this paper, we would like to thank the following individuals from Tencent for their contributions to the multilingual sandbox and manual annotation work: Hebin Li, Jinxu Hu, Bin Li, Zhihua Xu, Yunqing Sun, Xian Wu, Xiaohan Lu, Bingxian Liu, Bo Li, Bo Song, Cheng Zhang, Wenqi Xie, Zhirong Zheng.

\clearpage
\bibliography{biblio}
\bibliographystyle{colm2024_conference}

\clearpage
\appendix
\section{Data Category and Language Distribution Statistics}
\label{appendix:a}

We prompt \texttt{Claude Sonnet 4} to generate 20 language-agnostic category labels for classification:

\begin{itemize}
  \item \textbf{Core Programming Concepts}: \textit{Language Fundamentals}, \textit{Functions \& Modules}, \textit{Object-Oriented Programming}, \textit{Functional Programming}, \textit{Memory Management \& Performance}, \textit{Error Handling \& Debugging}
  
  \item \textbf{Data and Algorithms}: \textit{Data Structures \& Collections}, \textit{Algorithms \& Problem Solving}, \textit{String \& Text Processing}, \textit{File \& I/O Operations}, \textit{Concurrency \& Async Programming}
  
  \item \textbf{Application Domains}: \textit{Network Programming \& Communication}, \textit{Database Operations \& Persistence}, \textit{Web Development \& Frameworks}, \textit{Mobile \& Cross-platform Development}, \textit{Systems Programming \& Low-level Development}
  
  \item \textbf{Advanced Topics and Tooling}: \textit{Data Science \& Analytics}, \textit{Machine Learning \& AI}, \textit{Testing \& Quality Assurance}, \textit{Development Tools \& Ecosystem}
\end{itemize}

\begin{figure}[t]
\centering
\begin{subfigure}[b]{0.24\textwidth}
    \centering
    \includegraphics[width=\textwidth]{figures/tag_acb.pdf}
    \caption{AutoCodeBench}
\end{subfigure}
\hfill
\begin{subfigure}[b]{0.24\textwidth}
    \centering
    \includegraphics[width=\textwidth]{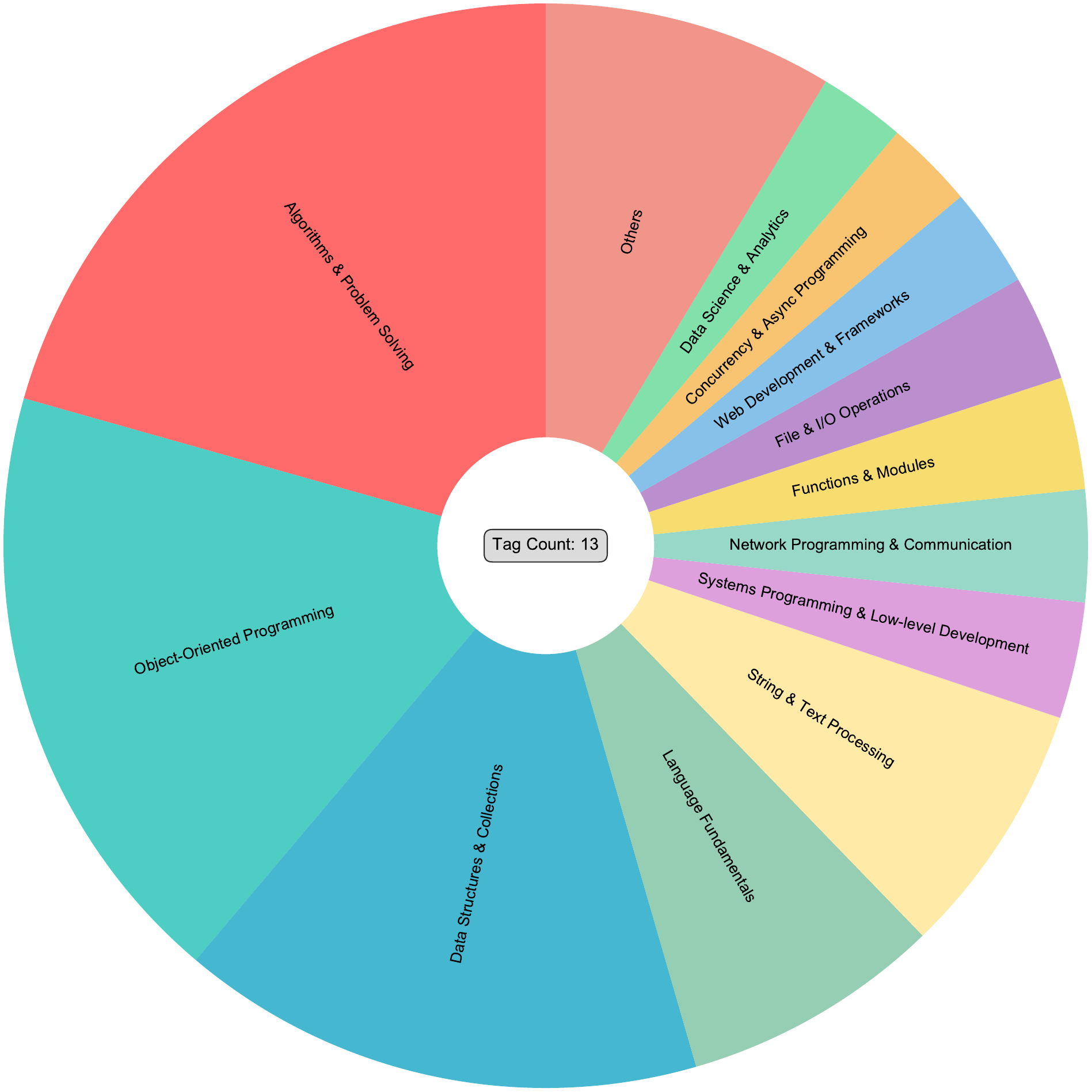}
    \caption{AutoCodeBench-Lite}
\end{subfigure}
\hfill
\begin{subfigure}[b]{0.24\textwidth}
    \centering
    \includegraphics[width=\textwidth]{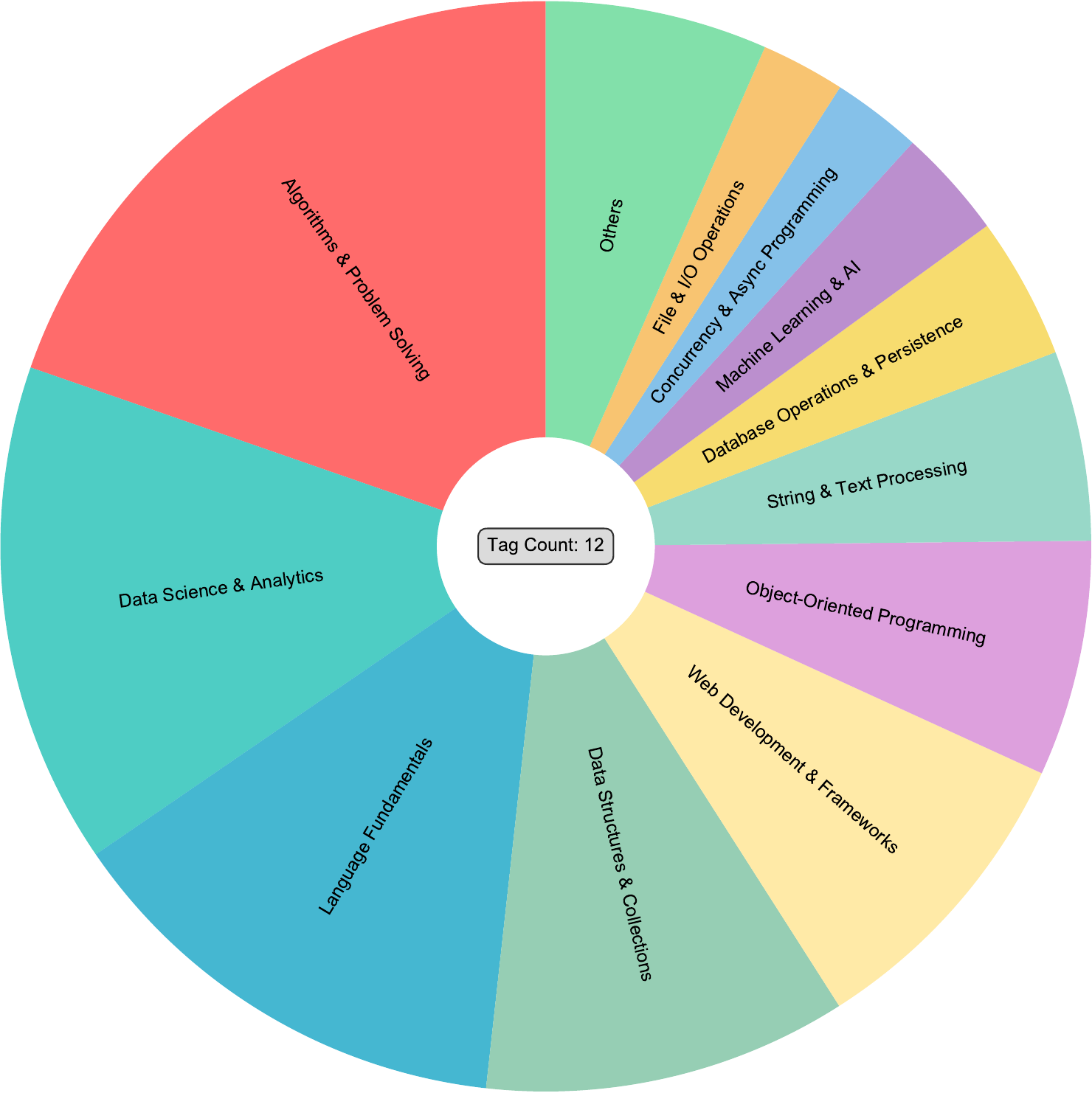}
    \caption{FullStackBench}
\end{subfigure}
\hfill
\begin{subfigure}[b]{0.24\textwidth}
    \centering
    \includegraphics[width=\textwidth]{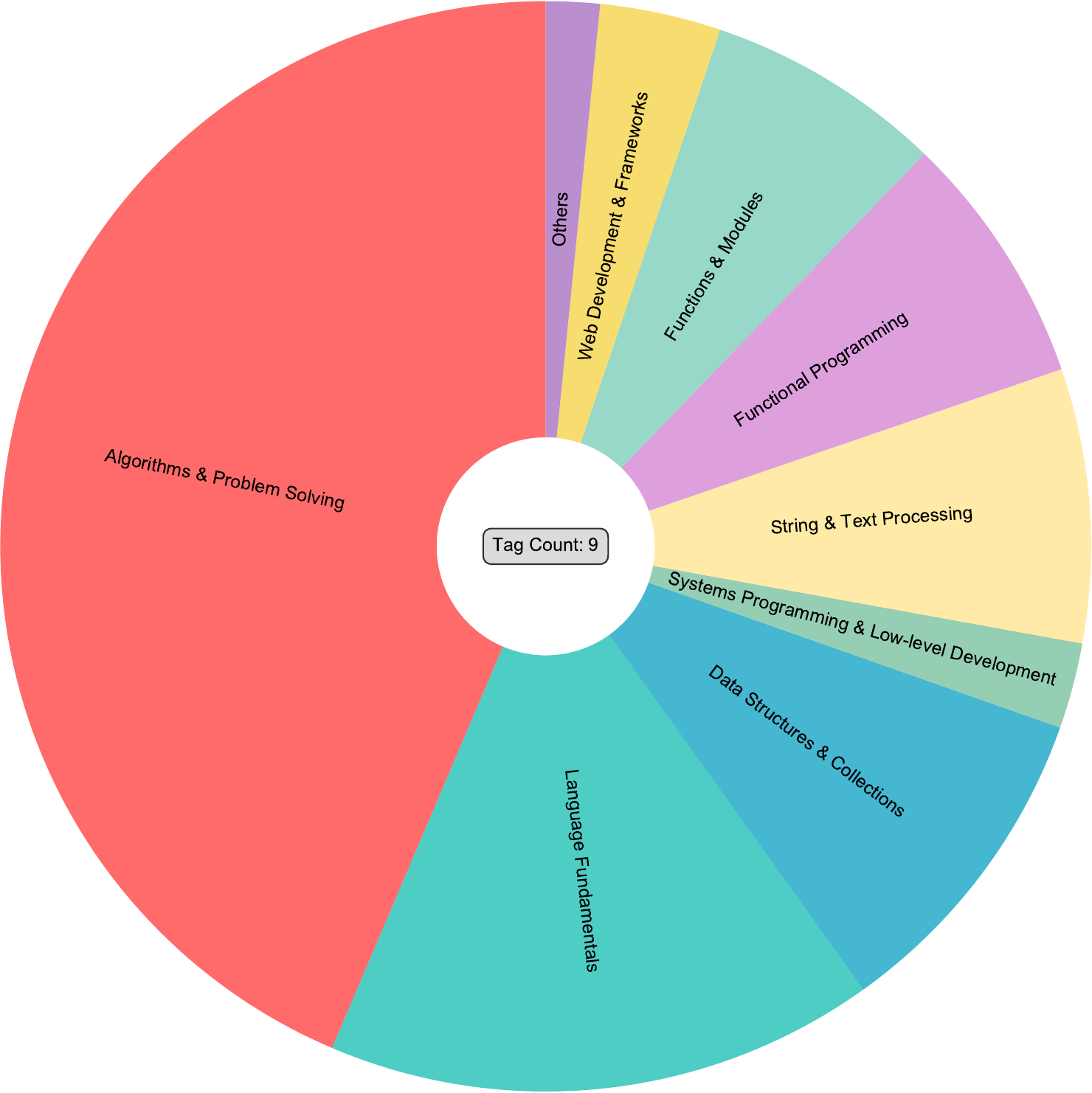}
    \caption{McEval}
\end{subfigure}

\caption{Category Distribution of Different Benchmarks.}
\label{appendix:tag}
\end{figure}

\begin{figure}[t]
\centering
\begin{subfigure}[b]{0.24\textwidth}
    \centering
    \includegraphics[width=\textwidth]{figures/lan_acb.pdf}
    \caption{AutoCodeBench}
\end{subfigure}
\hfill
\begin{subfigure}[b]{0.24\textwidth}
    \centering
    \includegraphics[width=\textwidth]{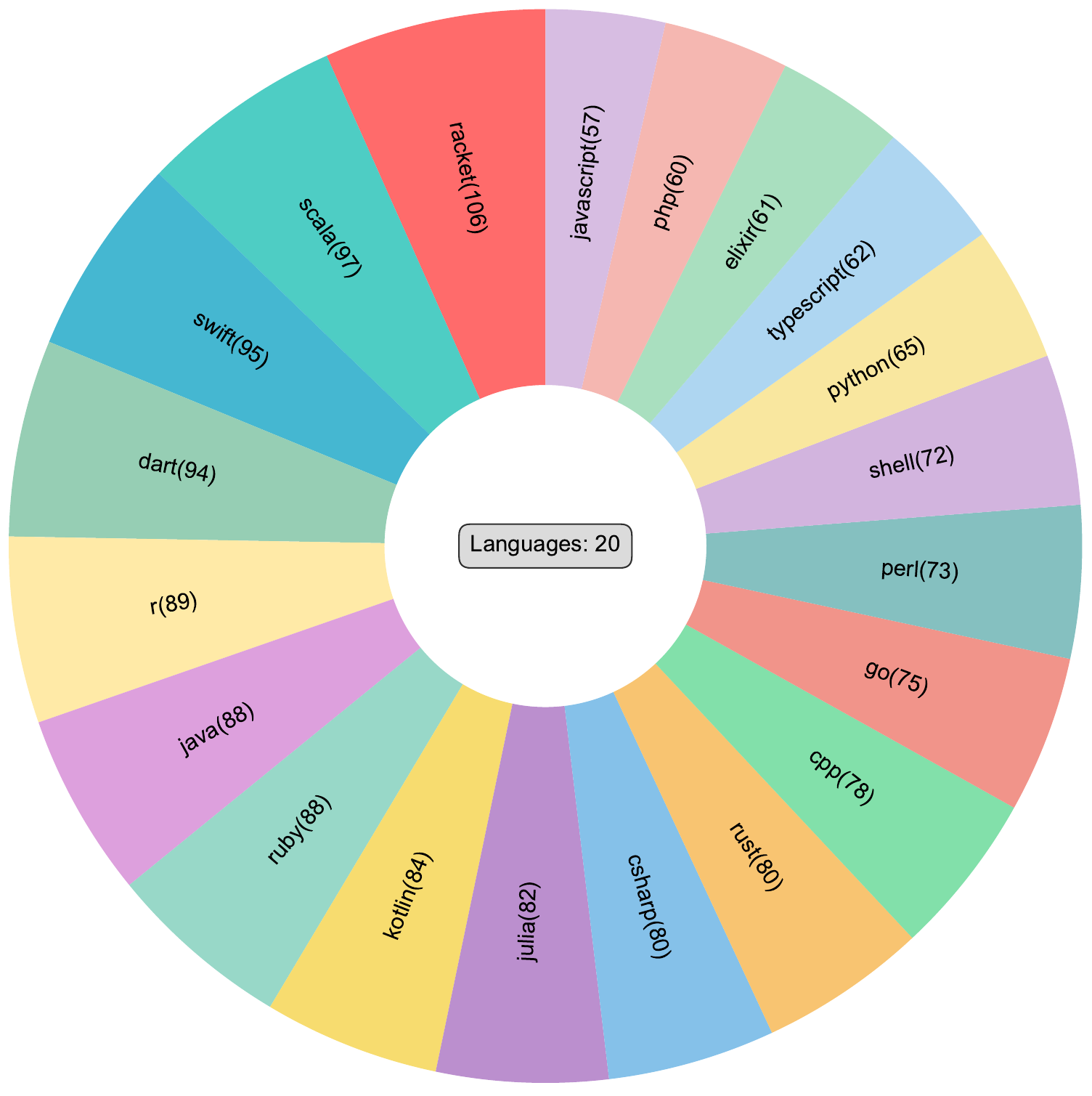}
    \caption{AutoCodeBench-Lite}
\end{subfigure}
\hfill
\begin{subfigure}[b]{0.24\textwidth}
    \centering
    \includegraphics[width=\textwidth]{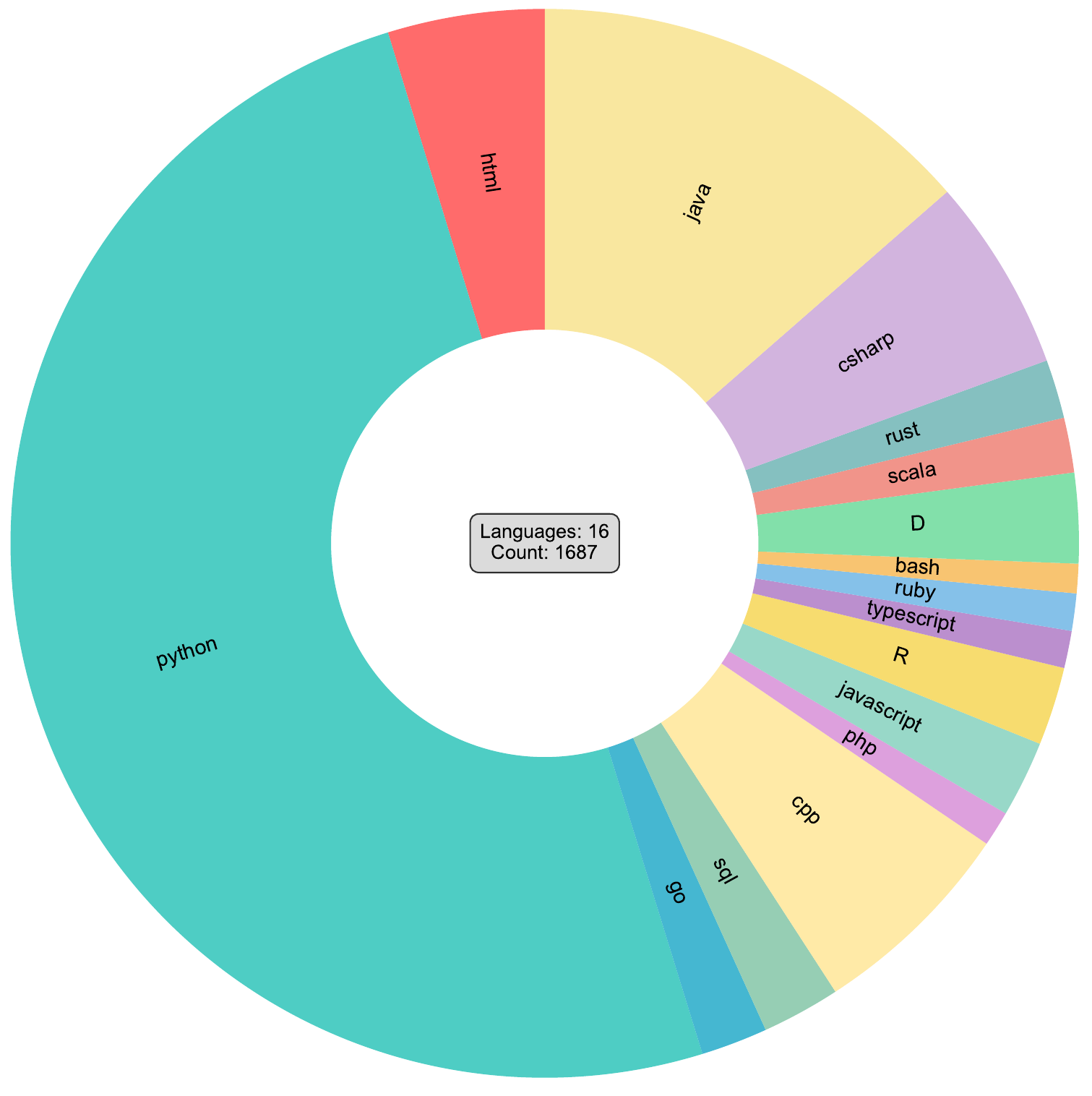}
    \caption{FullStackBench}
\end{subfigure}
\hfill
\begin{subfigure}[b]{0.24\textwidth}
    \centering
    \includegraphics[width=\textwidth]{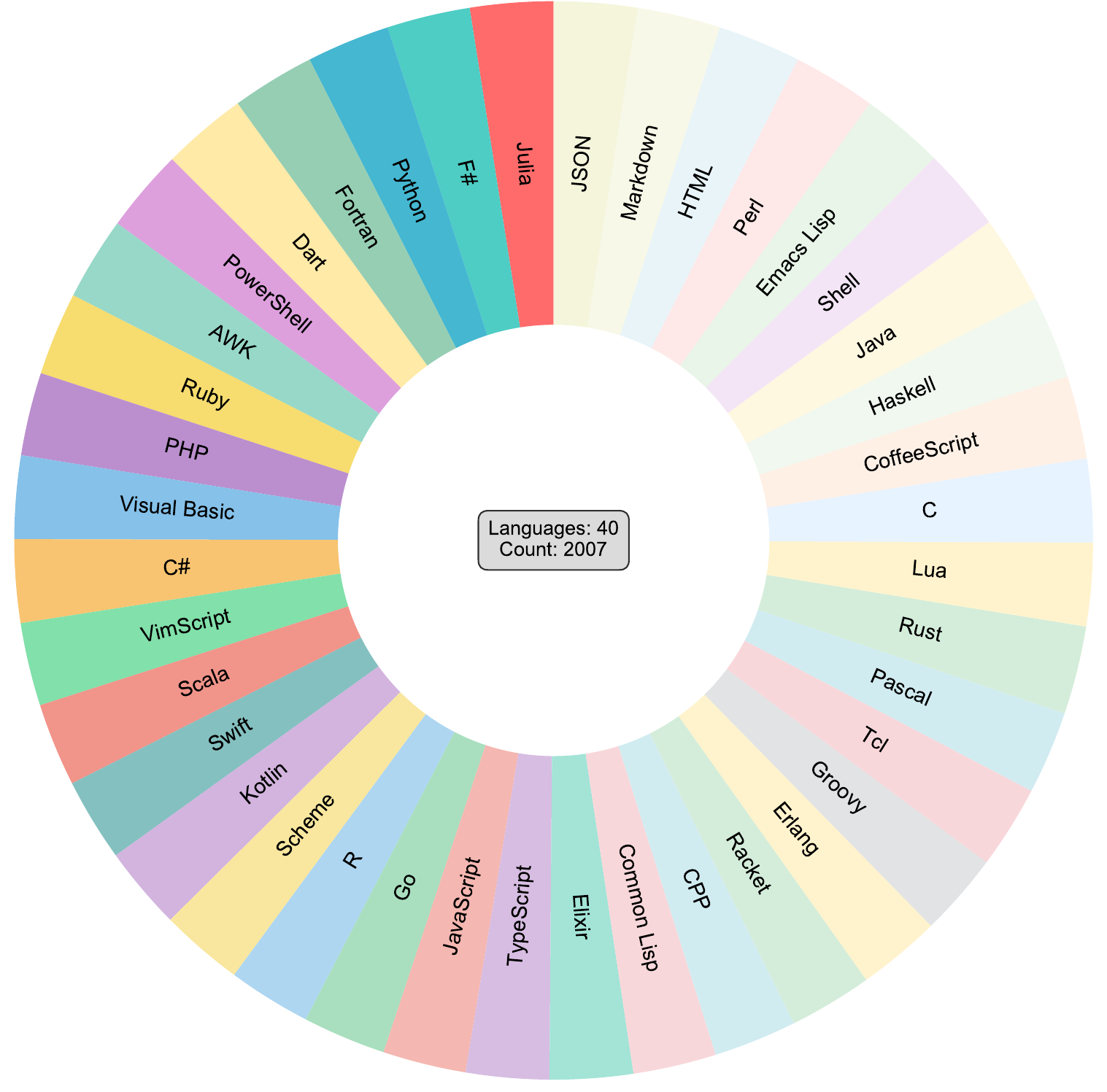}
    \caption{McEval}
\end{subfigure}

\caption{Language Distribution of Different Benchmarks.}
\label{appendix:lan}
\end{figure}

In addition to AutoCodeBench, we conduct task tagging and language distribution analysis for AutoCodeBench-Lite, FullStackBench, and McEval. The results are presented in Figures~\ref{appendix:tag} and~\ref{appendix:lan}. FullStackBench demonstrates comparable category diversity to AutoCodeBench(-Lite) but suffers from an imbalanced language distribution. In contrast, McEval exhibits a well-balanced multilingual distribution but lacks diversity and balance in its category coverage. Our AutoCodeBench(-Lite) achieves the most comprehensive category coverage while maintaining a balanced multilingual distribution, enabling thorough and accurate evaluation of LLMs' multilingual code generation capabilities.

\section{Manual Verification}
\label{appendix:b}

\begin{table}[t]
\small
\begin{tabular}{lccccccc}
\toprule
                    & \textbf{Average}  & \textbf{Python} & \textbf{C++} & \textbf{Java} & \textbf{JS} & \textbf{Go} & \textbf{Shell} \\
                    \midrule
Problem Accuracy    & 87.6 & 83.5            & 88.0           & 86.0            & 89.0          & 86.0          & 93.3           \\
\midrule
Current Upper Bound & 66.9\textcolor{red}{$_{\Delta20.7}$} & 61.7\textcolor{red}{$_{\Delta21.8}$}            & 71.5\textcolor{red}{$_{\Delta16.5}$}         & 76.1\textcolor{red}{$_{\Delta9.9}$}          & 58.2\textcolor{red}{$_{\Delta30.8}$}        & 65.4\textcolor{red}{$_{\Delta20.6}$}        & 68.6\textcolor{red}{$_{\Delta24.7}$}           \\
\midrule
Claude 4 Opus (Reasoning)    & 44.6\textcolor{red}{$_{\Delta43.0}$} & 40.3\textcolor{red}{$_{\Delta43.2}$}            & 44.1\textcolor{red}{$_{\Delta43.9}$}         & 55.9\textcolor{red}{$_{\Delta30.1}$}          & 38.6\textcolor{red}{$_{\Delta50.4}$}        & 37.2\textcolor{red}{$_{\Delta48.8}$}        & 51.6\textcolor{red}{$_{\Delta41.7}$}   \\
\bottomrule
\end{tabular}
\caption{Comparison of Accuracy, Upper Bound, and Model Performance.}
\label{appendix:accuracy}
\end{table}

\begin{figure}[t]
\centering
\includegraphics[width=0.9\textwidth]{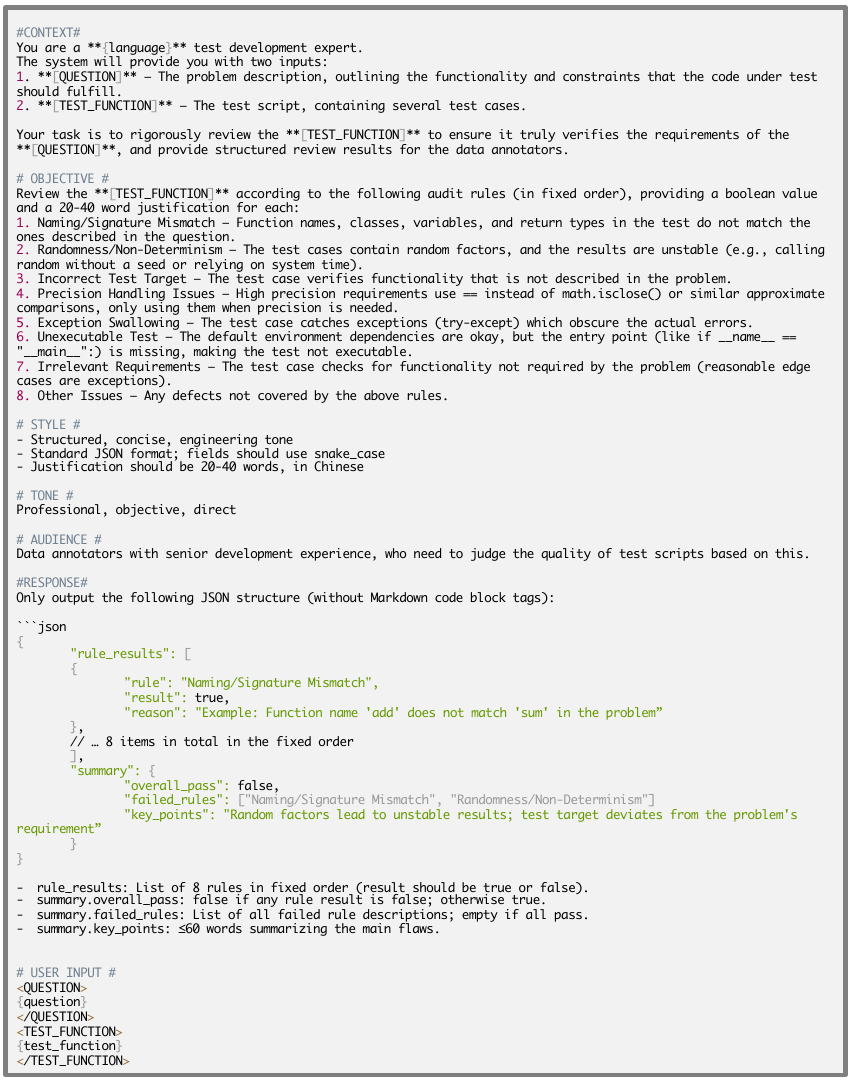}
\caption{The English prompt of annotation and critic.}
\label{appendix:prompt_annotation}
\end{figure}

Due to the involvement of multiple languages and domains in the data, directly verifying the quality of the data through manual annotation presents significant challenges. To address this issue, we employ a Human-LLM collaboration approach for data quality validation. Specifically, we design prompts in the native languages of the annotators and use the \texttt{DeepSeek-R1-0528} to generate detailed reasoning processes and checklist-based annotation results. The prompt is shown in Figure~\ref{appendix:prompt_annotation}. During the annotation process, we assume that the programming problems are completely correct. The primary task of the annotators is to assess the correctness of the test functions and their alignment with the programming problem, based on the LLM's output. We allow for test cases that may not cover all boundary conditions, focusing primarily on the correctness of the test functions rather than their comprehensiveness. The annotators pay particular attention to the following aspects:
\begin{itemize}
    \item Whether the function names, class names, variable definitions, and return types are consistent with the problem description;
    \item Whether the test cases exhibit randomness or non-reproducibility;
    \item Whether the test cases contradict the logic presented in the problem statement;
    \item Whether there are any precision issues with the test cases;
    \item Whether the test functions include test cases that are not addressed in the problem description.
\end{itemize}


To facilitate annotation, we design a front-end interface, including question, test function, critic reasoning process and results from LLM. Based on the information provided in the interface, the annotators generate a binary classification output.

Using the annotation results, we calculate the problem accuracy rates for different programming languages (Python, C++, Java, JavaScript, Go, Shell), as shown in Table~\ref{appendix:accuracy}. The results indicate that, despite the presence of some noisy data, our benchmark model still demonstrates high accuracy (87.6\%). Furthermore, even after removing the noise, the current SOTA model shows significant room for improvement ($\Delta43.0$), further validating the high difficulty level of our benchmark. Besides, we find that, compared to logic errors in the problem description and errors in the test functions, the most frequently occurring issue is \textbf{incomplete problem descriptions}. For example, some test functions reference class or function names that are essential but not explicitly mentioned in the problem description, or they require natural language outputs for edge cases that are not explicitly specified in the problem statement, leading to mismatches between the generated code and the test functions. Interestingly, we observe similar issues in manually annotated BigCodeBench~\citep{zhuo2025bigcodebench}, highlighting the significant challenge of creating comprehensive and accurate programming problems for annotators.

\begin{figure*}[t]
    \centering
\includegraphics[width=1.0\columnwidth]{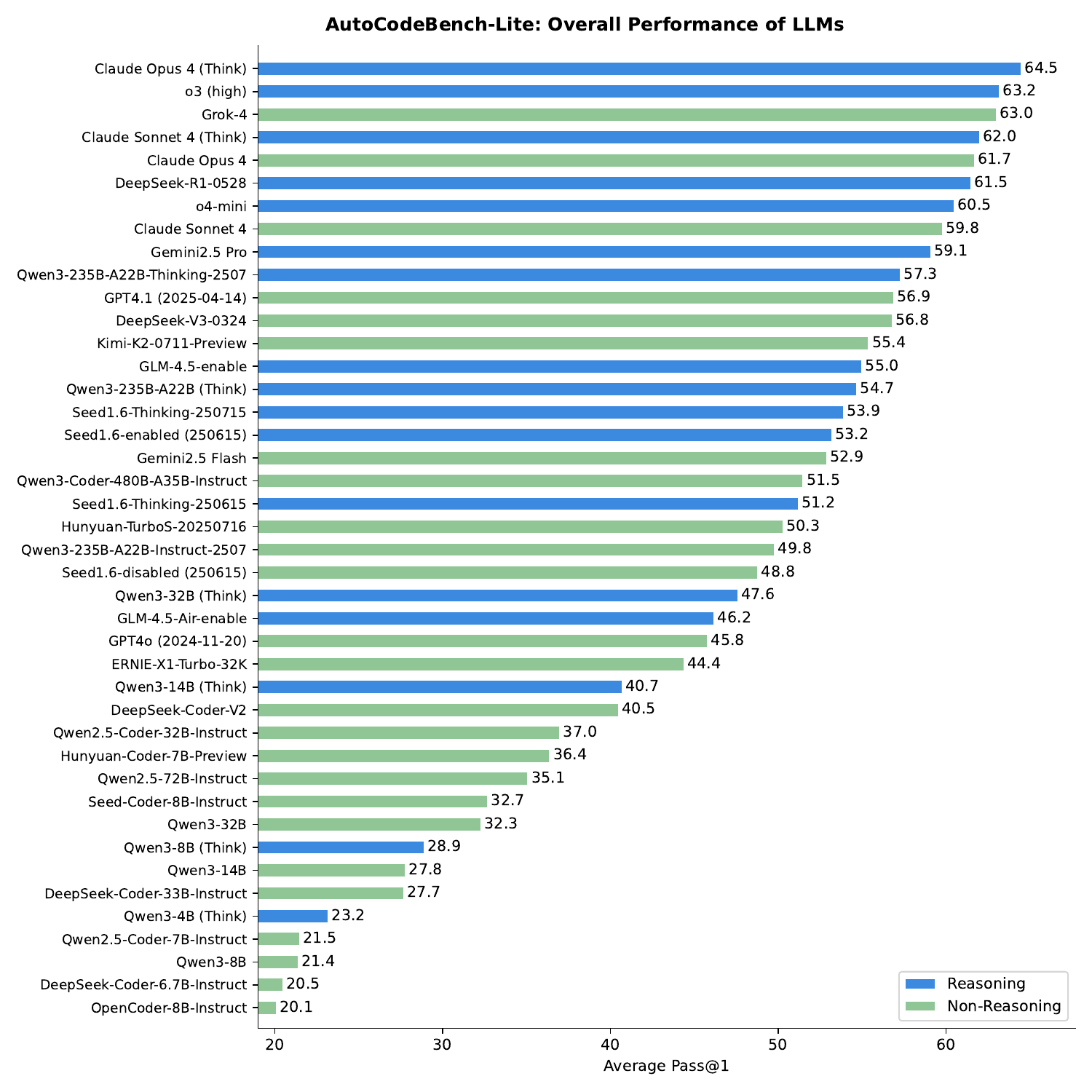}
    \caption{AutoCodeBench-Lite leaderboard showing Pass@1 performance of various LLMs.}
	\label{appendix:leaderboard_lite}
\end{figure*}

\section{Multilingual Code Sandbox Service}

This service offers a secure and high-performance environment for the compilation and execution of code in over 30 programming languages. It supports large-scale code data validation, making it suitable for high-volume, automated testing scenarios. Our multilingual sandbox has the following features:

\begin{itemize}
    \item \textbf{Multilingual Support}: The service supports more than 30 programming languages, including popular ones like Python, JavaScript, Go, Java, C++, and Rust, providing versatility for various use cases.
    \item \textbf{Security Isolation}: Code execution is isolated within Docker containers, ensuring that each execution environment is separate. Additionally, iptables firewall rules are applied to maintain a high level of security, preventing unauthorized access or interference.
    \item \textbf{Smart Code Integration}: The system automatically manages the integration of function code with testing code. It adapts to language-specific syntax, ensuring seamless execution without requiring manual intervention for code merging.
    \item \textbf{High Performance}: Powered by a Gunicorn multi-process architecture, the sandbox supports concurrent execution of multiple code instances, making it capable of handling a high volume of requests efficiently.
    \item \textbf{RESTful API}: The service provides a clean and easy-to-use HTTP-based API, allowing developers to interact with the sandbox programmatically, whether for integrating into larger applications or automating tasks.
    \item \textbf{Extensive Language Support}: Beyond the mainstream languages, the sandbox also supports emerging and niche languages, allowing it to cater to a wide variety of development environments and user needs.
    \item \textbf{Custom Execution Environments}: Users can configure specific environments for their tasks, enabling tailored execution conditions based on their unique requirements.
\end{itemize}

\section{Prompts for Automated Workflow}

\begin{figure}[t]
\centering
\includegraphics[width=0.9\textwidth]{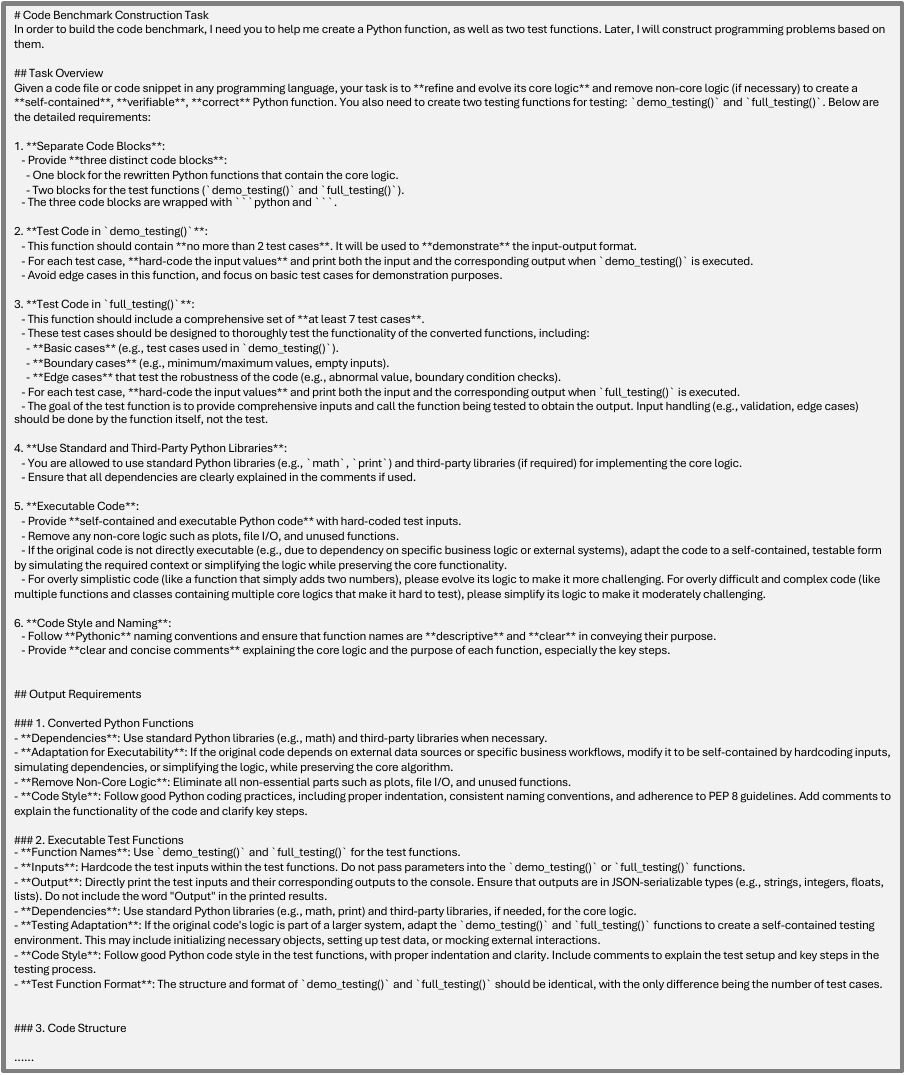}
\caption{The prompt of generating code solution. Due to the excessive length of the prompt, we have omitted the latter part.}
\label{appendix:prompt_solution}
\end{figure}

The prompt of generating code solution is shown in Figure~\ref{appendix:prompt_solution}.

\begin{figure}[t]
\centering
\includegraphics[width=0.9\textwidth]{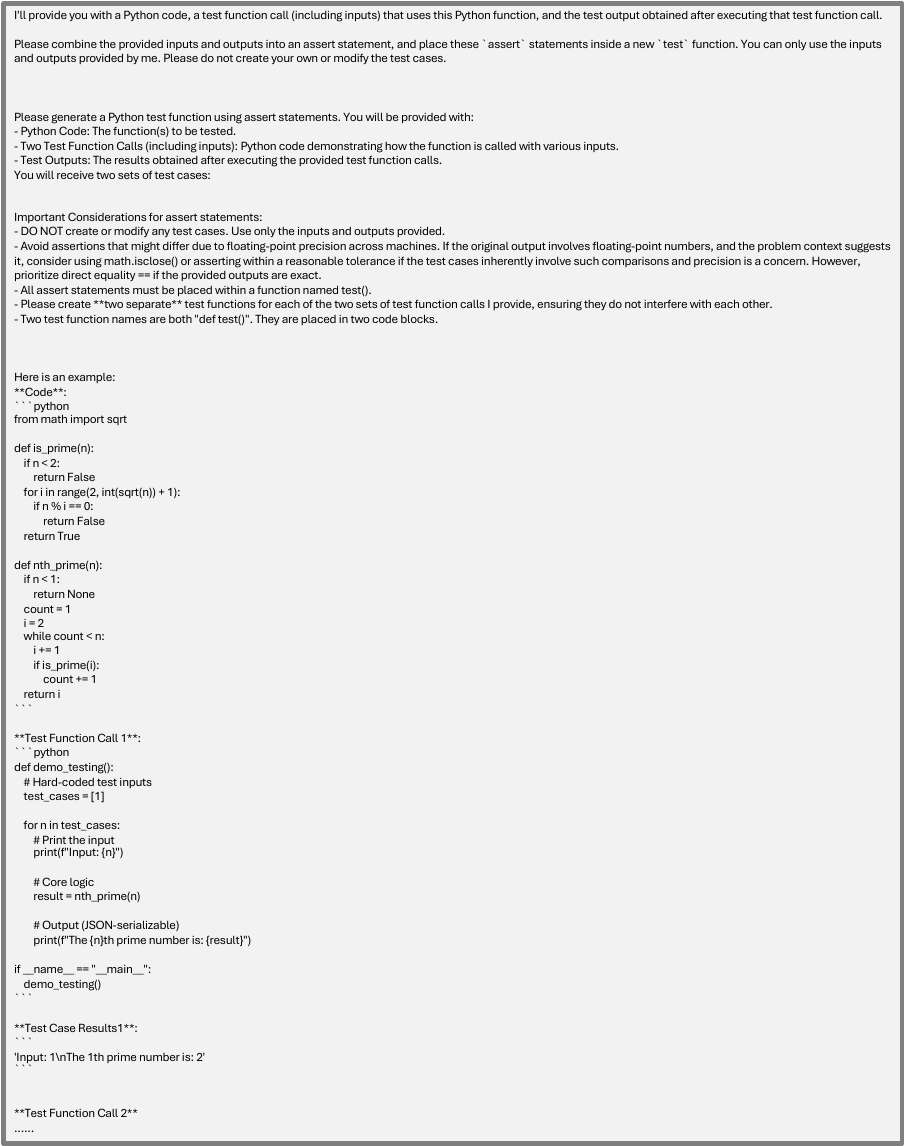}
\caption{The prompt of generating test function. Due to the excessive length of the prompt, we have omitted the latter part.}
\label{appendix:prompt_test}
\end{figure}

The prompt of generating test function is shown in Figure~\ref{appendix:prompt_test}.

\begin{figure}[t]
\centering
\includegraphics[width=0.9\textwidth]{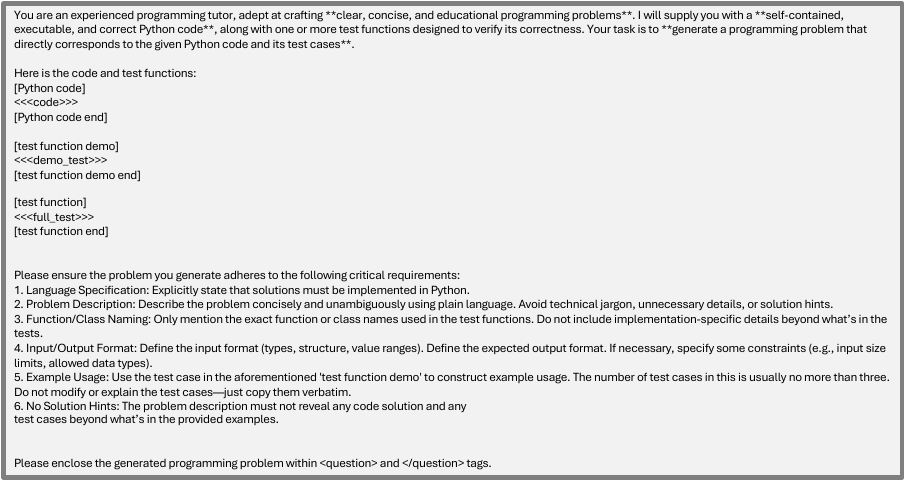}
\caption{The prompt of generating programming problem.}
\label{appendix:prompt_problem}
\end{figure}

The prompt of generating programming problem is shown in Figure~\ref{appendix:prompt_problem}.

The prompt of LLM-as-Critic is shown in Figure~\ref{appendix:prompt_annotation}.

\begin{figure}[t]
\centering
\includegraphics[width=0.9\textwidth]{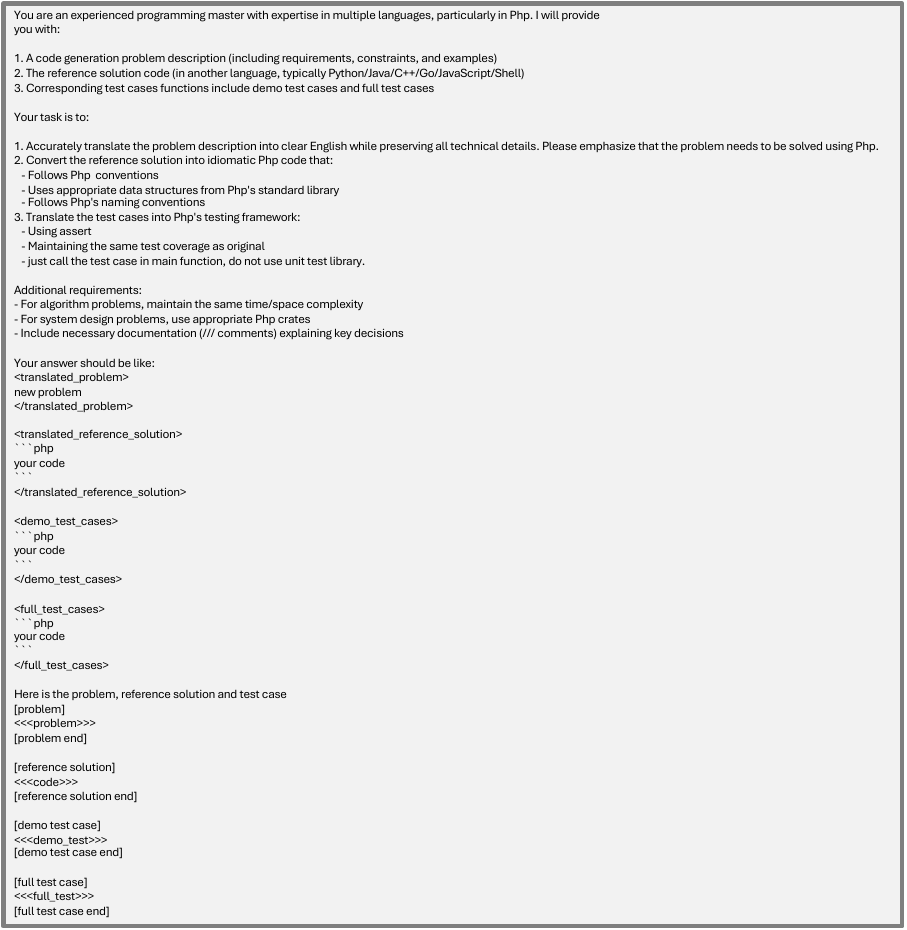}
\caption{The prompt of translating languages.}
\label{appendix:prompt_translate}
\end{figure}

The prompt of translating languages is shown in Figure~\ref{appendix:prompt_translate}.

\end{document}